\documentclass[10pt,twocolumn,letterpaper]{article}

\usepackage{cvpr}
\usepackage{times}
\usepackage{epsfig}
\usepackage{epstopdf}
\usepackage{graphicx}
\usepackage{amsmath}
\usepackage{amssymb}
\usepackage{subcaption}
\usepackage{multirow}
\usepackage[export]{adjustbox}
\usepackage{array}
\usepackage[tableposition=bottom]{caption}

\usepackage[pagebackref=true,breaklinks=true,letterpaper=true,colorlinks,bookmarks=false]{hyperref}

\cvprfinalcopy

\begin{document}

\title{MaskFace: multi-task face and landmark detector}

\author{Dmitry Yashunin, Tamir Baydasov, Roman Vlasov\\
Harman\\
{\tt\small \{Dmitry.Yashunin, Tamir.Baydasov, Roman.Vlasov\}@harman.com}
}

\maketitle

\begin{abstract}
Currently in the domain of facial analysis single task approaches for face detection and landmark localization dominate. In this paper we draw attention to multi-task models solving both tasks simultaneously. We present a highly accurate model for face and landmark detection. The method, called MaskFace, extends previous face detection approaches by adding a keypoint prediction head. The new keypoint head adopts ideas of Mask R-CNN by extracting facial features with a RoIAlign layer. The keypoint head adds small computational overhead in the case of few faces in the image while improving the accuracy dramatically. We evaluate MaskFace's performance on a face detection task on the AFW, PASCAL face, FDDB, WIDER FACE datasets and a landmark localization task on the AFLW, 300W datasets. For both tasks MaskFace achieves state-of-the-art results outperforming many of single-task and multi-task models.
\end{abstract}

\section{Introduction}

In recent years facial image analysis tasks became very popular because of their attractive practical applications in automotive, security, retail and social networks. Face analysis starts from the basic tasks of bounding box detection and landmark localization. The common pipeline is to sequentially apply single-task models to solve these problems independently: 1) detect faces, 2) detect landmarks (also called keypoints)~\cite{wang2018deep}.

However, it is a challenge to develop a software system consisting of many sequentially applied convolutional neural networks (CNN) because each CNN should be trained separately and deals with errors made by previous models. Different heuristics and special training procedures are applied to achieve robustness of the overall system. Crucially, single-task CNNs can't benefit from shared deep representations and additional supervision provided by multiple tasks. Recent studies show that multi-task CNNs that produce multiple predictive outputs can offer higher accuracy and better speed than single-task counterparts but are difficult to train properly \cite{he2017mask,chen2017gradnorm,maninis2019attentive}.

\begin{figure}[t]
   \centering
   \includegraphics[width=1\linewidth,trim={0cm 0cm 0cm 0cm}]{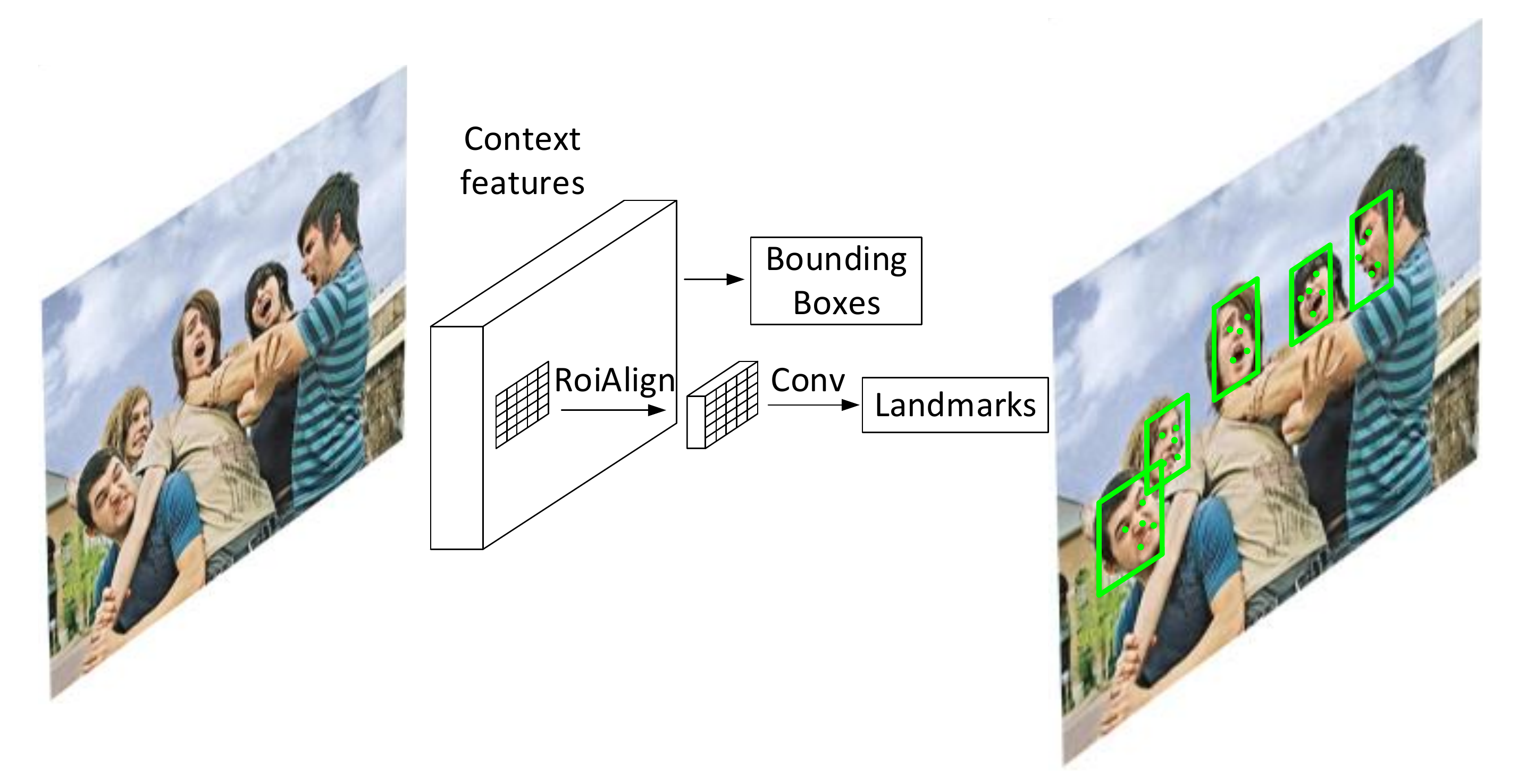}
   \caption{MaskFace design. Predicted bounding boxes from context features are directly used for feature extraction and facial landmark localization.}
\label{fig:MaskFace_intro}
\end{figure}

\begin{figure*}[t]
   \centering
   \includegraphics[width=0.8\textwidth,trim={20cm 10cm 16cm 6cm}]{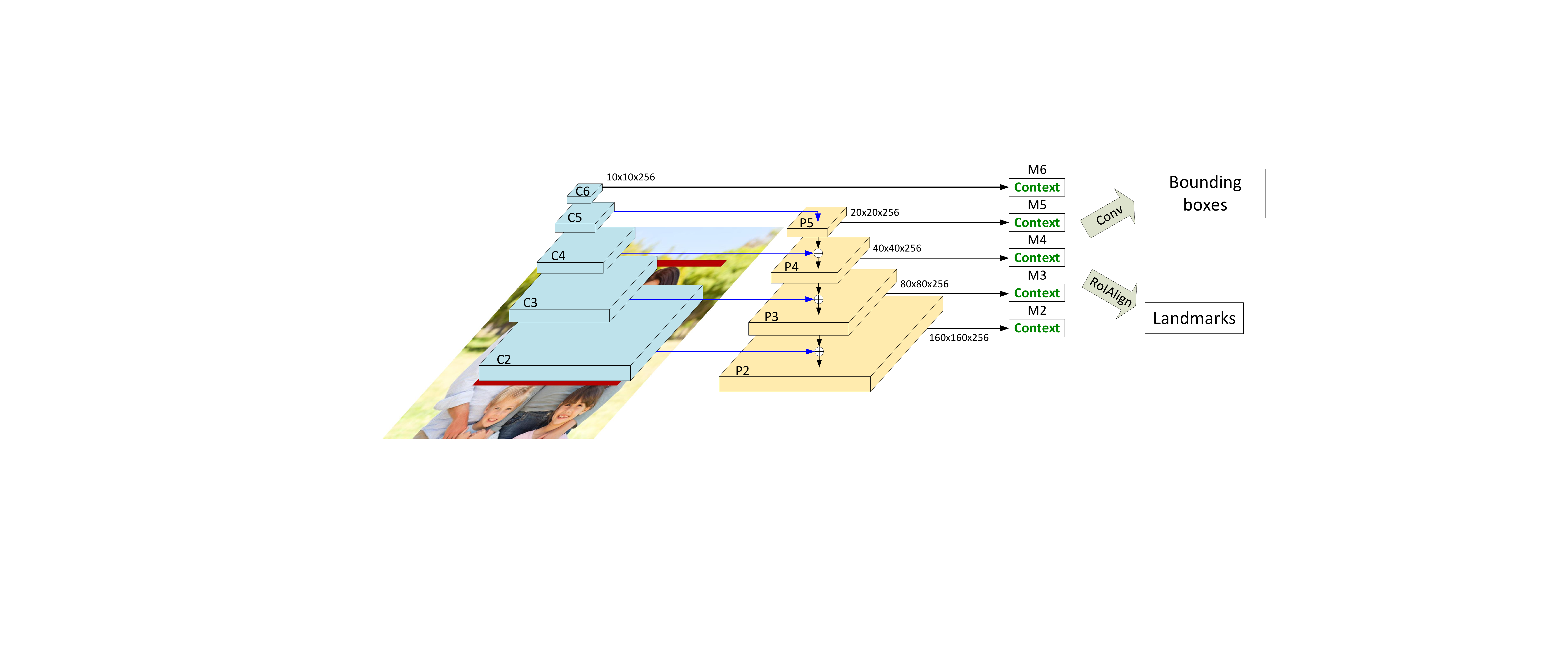}
   \caption{Outline of the proposed approach. MaskFace adopts the feature pyramid followed by independent context modules. Outputs of context modules are used for face detection and landmark localization.}
\label{fig:Model}
\end{figure*}

We argue that despite the recent achievements in multi-task models in the domain of facial analysis they receive low attention and their accuracy loses to single-task rivals. The most popular multi-task model MTCNN uses cascades of shallow CNNs, but they do not share feature representations \cite{zhang2016joint,cai2018joint}. Modern end-to-end multi-task approaches are mainly represented by single-shot methods. For landmark localization either regression heads \cite{chen2018real,deng2019retinaface} or heatmaps of keypoints \cite{wang2019coupled,liu2019facial} are used. The heatmap based approaches suffer from low face detection accuracy, while regression-based ones have worse landmark localization. The reason is that regression-based methods can't afford strong landmark prediction heads. In addition, there is a misalignment between the spatially discrete features of activation maps and continuous positions of facial landmarks. The misalignment can't be properly handled by shallow convolutional layers.

In this paper we propose an accurate multi-task face and landmark detection model that combines advantages of previous approaches. Our model extends popular face detection approaches such as RetinaFace \cite{deng2019retinaface} and SSH \cite{najibi2017ssh} by adopting ideas of Mask R-CNN \cite{he2017mask}. At the first stage we predict bounding boxes, at the second stage predicted bounding boxes are used for extraction of facial features from shared representations (see Figure \ref{fig:MaskFace_intro}). Unlike Mask R-CNN and other multi-stage approaches we predict bounding boxes in a single forward pass, that allows to increase performance \cite{he2017mask,ren2015faster,lin2017feature,lin2017focal}. For feature extraction we use a RoIAlign layer \cite{he2017mask} offering good pixel-to-pixel alignment between predicted bounding boxes and discrete feature maps. To improve detection of tiny faces we use a feature pyramid \cite{lin2017feature} and context modules \cite{deng2019retinaface,najibi2017ssh,earp2019face} (see Figure \ref{fig:Model}, ~\ref{fig:Context}). The feature pyramid transmits deep features to shallow layers, while the context modules increase a receptive field and make prediction layers stronger. The MaskFace's landmark head is as fast as the original Mask R-CNN head. In the case of few faces in the image prediction of landmarks adds negligible computational overhead.

In summary our contribution is twofold: Firstly, we propose using the mask head for facial keypoint detection and we perform experiments on sparse as well as dense landmarks that are usually omitted in previous multi-task models. And secondly, we show how using the mask head we achieve state-of-the-art results on both the face detection and landmark localization tasks. We systematically study how different model parameters influence the accuracy and perform extensive experiments on popular datasets. This highlights that a well-designed model has a significant impact on the result and can outperform many of more sophisticated approaches.

\section{Related work}

Object detectors are mainly based on the idea of default bounding boxes (also called anchors) that densely cover an image at all scales \cite{ren2015faster}. Anchors are classified, and their positions are refined. In single-stage face detectors bounding boxes are predicted in a single forward pass \cite{liu2016ssd,zhang2017faceboxes}, in two-stage detectors there are two rounds of anchor refinement \cite{ren2015faster,li2019dsfd}. Anchor-based approaches have the significant class imbalance between positive and negative anchors. The class imbalance problem is usually solved by online hard element mining (OHEM) \cite{liu2016ssd} or a dynamically scaled cross entropy loss (focal loss) \cite{lin2017focal,liu2016ssd}.

To detect difficult objects, such as tiny faces or faces with complex poses and high occlusion, different approaches are applied. In \cite{zhang2017faceboxes,li2019dsfd} the authors use densified anchor tiling. In Feature Pyramid Network (FPN) \cite{lin2017feature} semantically strong features with low resolution are combined with weak features from high resolution layers. Also, to improve detection of small objects context is incorporated. In two-stage detectors, context can be modeled by explicitly enlarging the window around the candidate proposals \cite{wu2019enhanced}. In single-shot methods context is incorporated by enlarging a receptive field by additional convolutional layers \cite{najibi2017ssh,li2019dsfd}.

Regression and segmentation methods are used for facial landmark prediction. Regression approaches are often based on $L_1$ and $L_2$ losses or their modifications \cite{feng2018wing,honari2018improving,barron2019general}. Also, multiple-stages of landmark refinement can be used to improve the accuracy \cite{zhu2016unconstrained,feng2018wing}. There are approaches that use 3D face reconstruction to predict dense landmarks \cite{feng2018joint,zhou2019dense}.

Multi-task models combine several-single task methods in one model. In MTCNN the authors use the image pyramid and cascades of shallow CNNs to predict face bounding boxes and landmarks. Recent methods adopt feature pyramids that naturally exist in CNNs \cite{lin2017feature}. For landmark localization additional regression heads \cite{chen2018real,deng2019retinaface} are added to commonly used detectors such as SSD \cite{liu2016ssd}, RetinaNet \cite{lin2017focal}. In \cite{deng2019retinaface,chaudhuri2019joint} the authors add branches that predict 3D shapes of faces. Mask R-CNN offers a general and flexible architecture for multi-tasking \cite{he2017mask}. It is based on a RoIAlign pooling that allows to extract features for proposals containing objects. Extracted features can be used for different tasks, for example, for segmentation, keypoint localization \cite{he2017mask} or predicting 3D shape of objects \cite{gkioxari2019mesh}.

Differences from the closest works. In difference to SHH, RetinaNet and other single and multi-task face detection models we add the mask head that significantly improves landmark localization accuracy. In the most papers devoted to face detection OHEM is used, while we show that a focal loss can offer state-of-the-art results. In difference to Mask R-CNN \cite{he2017mask} we predict bounding boxes in a single forward pass. In difference to RetinaMask \cite{fu2019retinamask} we use context modules and focus on the face detection and landmark localization.

\section{Method}

\subsection{Multi-task loss}

We model a landmark location as a one-hot mask, and adopt MaskFace to predict $K$ masks, one for each of the $K$ landmarks, e.g. left eye, right eye, etc. The multi-task loss for an image is defined as:
\begin{equation}\label{first_eq}
L=L_{cls}+L_{box}+ \lambda_{kp} L_{kp},
\end{equation}
where $L_{cls}$ is an anchor binary classification loss (face vs background),
$L_{box}$ is a regression loss of anchors’ positions,
$L_{kp}$ is a localization loss of keypoints weighted with a parameter $\lambda_{kp}$.

For anchor classification we use a focal loss \cite{lin2017focal}:
\begin{equation}\label{secnd_eq}
\begin{split}
L_{cls}= -\frac{1}{N_{pos}}*[ \alpha\sum_{i \in Pos}(1-p_i)^\gamma \log p_i \\
+ (1 - \alpha)\sum_{i \in Neg}p_i^\gamma \log(1-p_i)].
\end{split}
\end{equation}

For bounding box regression we apply a smooth version of $L_1$ loss:
\begin{equation}\label{third_eq}
L_{box}= \frac{1}{N_{pos}} \sum_{i \in Pos}\mbox{smooth}_{\mbox{L1}}(t_i-t_i^*). 
\end{equation}

To predict landmark locations, we apply a spatial cross-entropy loss to each of the landmark masks:
\begin{equation}\label{forth_eq}
L_{kp} = -\frac{1}{KN_{pos}}  \sum_{i \in Pos} \sum_{k=1}^K \log M_{i,k,j_{i,k}^*,l_{i,k}^*},
\end{equation}
\begin{equation}\label{fifth_eq}
M_{i,k,j,l} = \frac{\exp{(L_{i,k,j,l})}}{\sum_{j=1}^m \sum_{l=1}^m \exp{(L_{i,k,j,l})}},
\end{equation}
where $p_i$ is a predicted probability of anchor $i$ being a face, $N_{pos}$ is a number of positive anchors, that should be classified as faces ($p_i$ should be equal to 1), negative anchors are ones that should be classified as background ($p_i$ should be equal to 0). Pos and Neg are sets of indices of positive and negative anchors, respectively. $\alpha$ is a balancing parameter between the classification loss of positive and negative anchors, $\gamma$ is a focusing parameter that reduces the loss for well-classified anchors. $t_i$ is a vector representing the 4 parameterized coordinates of a predicted bounding box, and $t_i^*$ is that of the ground-truth box associated with a positive anchor $i$. $L_{i,k,j,l}$ and $M_{i,k,j,l}$ are predicted logits and a mask for a landmark $k$ for a positive sample $i$, respectively. $j_{i,k}^*$, $l_{i,k}^*$ are indices of mask pixels at which a ground truth landmark $k$ in a positive sample $i$ is located.
For each of the $K$ keypoints the training target is a one-hot $m*m$ binary mask where only a single pixel is labeled as foreground \cite{he2017mask}.

Parameters $\alpha$ and $\gamma$ are set to 0.25 and 2, respectively, following ~\cite{deng2019retinaface}. If not mentioned, we select a value of the keypoint loss weight $\lambda_{kp}$ equal to 0.25. The higher $\lambda_{kp}$ the more accurate localization of landmarks but face detection degrades. Our experiments show that $\lambda_{kp}$ = 0.25 gives a good trade-off between the accuracy of face detection and landmark localization.


\subsection{Architecture}

MaskFace design is straightforward and based on the \textit{maskrcnn-benchmark} \cite{massa2018mrcnn}. MaskFace has two prediction heads: face detection and landmark localization. The detection head outputs bounding boxes of faces. Predicted bounding boxes are used to extract face features from fine resolution layers allowing precise localization of landmarks. To achieve good pixel-to-pixel alignment during feature extraction we adopt a RoIAlign layer following Mask R-CNN \cite{he2017mask}. Extracted face features are used to predict localization masks of landmarks.

\noindent
\textbf{Face detection head.} We adopt the FPN architecture \cite{lin2017feature}. FPN combines low-resolution, semantically strong features with high-resolution, semantically weak features via a top-down pathway and lateral connections (see Figure \ref{fig:Model}). The result is a feature pyramid with rich semantics at all levels, that is necessary for detection of tiny faces. FPN outputs a set of feature maps called \{${P_2, P_3, P_4, P_5, P_6}$\} with 256 channels each and strides of \{$4, 8, 16, 32, 64$\}, respectively. Feature maps from $P_2$ to $P_5$ are calculated using feature maps of last layers with strides of \{$4, 8, 16, 32, 64$\}, respectively, that are called \{${C_2, C_3, C_4, C_5}$\}. \{${P_2, P_3, P_4, P_5}$\} layers have the same spatial size as the corresponding \{${C_2, C_3, C_4, C_5}$\} layers. $P_6$ is calculated by applying a max-pooling layer with a stride of 2 to $C_5$.

\begin{figure}[h]
   \centering
   \includegraphics[width=0.5\linewidth,trim={3cm 0cm 3cm 0cm}]{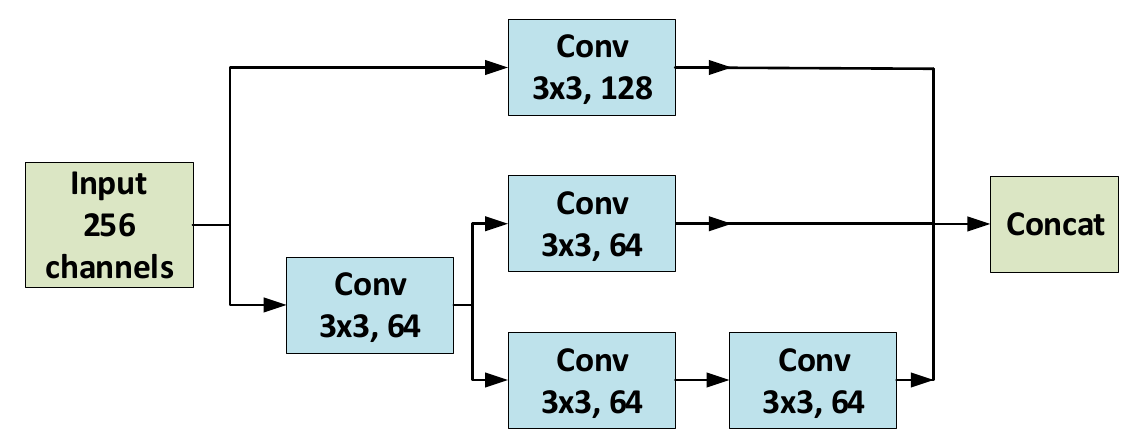}
   \caption{Context module design. Notation of convolutional layers: filter size, number of channels.}
\label{fig:Context}
\end{figure}

To increase the receptive field and add context to predictions we apply context modules with independent weights to \{${P_2, P_3, P_4, P_5, P_6}$\} \cite{najibi2017ssh,deng2019retinaface}. We call feature maps of context modules \{${M_2, M_3, M_4, M_5, M_6}$\}. The context module design is similar to the inception module (see Figure \ref{fig:Context}) \cite{szegedy2015going}. Sequential 3x3 convolutional filters are used instead of larger convolutional filters to reduce the number of calculations. ReLU is applied after each convolutional layer. Outputs of all branches are concatenated. Our experiments suggest that the context modules improve accuracy.

Feature maps after context modules are used for anchor boxes’ regression and classification by applying 1x1 convolutional layers with shared weights. Note that unlike other multi-stage detectors we do not use a second stage of bounding box refinement to increase performance.

We use translation-invariant anchor boxes similar to \cite{ren2015faster}. The base anchors have areas of \{$16^2, 32^2, 64^2, 128^2, 256^2$\} on pyramid levels from $M^2$ to $M^6$, respectively. At each pyramid level we use anchors with sizes of \{$2^0, 2^\frac{1}{3}, 2^\frac{2}{3}$\} of the base anchors for dense scale coverage. All anchors have the same aspect ratio of 1.0. There are three anchors per level and across levels they cover the scale range 16 -- 406 pixels. In total there are around 112k anchors for an input image of 640x640 resolution.

Anchor matching. If an anchor box has an intersection-over-union (IoU) overlap with a ground truth box greater than 0.5 then the anchor is considered as a positive example. If the overlap is less than 0.3 the anchor is assigned a negative label. All anchors with overlaps between 0.3 and 0.5 are ignored during training. Additionally, for anchor assigning we use low-quality matching strategy. For each ground truth box, we find a set of anchor boxes that have the maximum overlap with it. For each anchor in the set, if the anchor is unmatched, then we match it to the ground truth with the highest IoU. Our experiments suggest using the low-quality matching strategy because it improves accuracy.

\noindent
\textbf{Landmark localization head.} Predictions from the detection head are treated as region-of-interest (RoI) to extract features for landmark localization. At first, proposals are filtered: predictions with confidences less than 0.02 are ignored, non-maximum suppression (NMS) ~\cite{bodla2017soft} with a threshold of 0.6 is applied to remaining predictions. After that proposals are matched with ground truth boxes. If an IoU overlap of proposals with ground truth boxes is higher than 0.6 then proposals are used for extracting landmark features from the appropriate layers of the feature pyramid \{${M_2, M_3, M_4, M_5, M_6}$\}.

Following FPN we assign a RoI of width $w_{roi}$ and height $h_{roi}$ to the level $M_k$ of our feature pyramid by:

\begin{equation}\label{sixth_eq}
k = max(2, \lfloor k_0 + \log_2 (\sqrt{w_{roi}h_{roi}}/224) \rfloor),
\end{equation}
where $k_0=4$. Eqn. \ref{sixth_eq} means that if an area of a predicted bounding box is smaller than $112^2$, it is assigned to the feature layer $M_2$, between $112^2$ to $224^2$ is assigned to $M_3$, etc. The maximum layer is $M_6$. Unlike the previous approaches ~\cite{lin2017feature,fu2019retinamask} we use the finer resolution feature map $M_2$ with a stride of 4 for feature extraction. Our experiments show that high resolution feature maps are essential for precise landmark localization. The lower $k_0$ the more precise landmark detection (see Section \ref{sec:ablation_experiments}). If not mentioned $k_0=3$ is used.

We adopt a RoIAlign layer \cite{he2017mask} to extract features from assigned feature maps. RoIAlign allows properly aligning of the extracted features with the input RoIs. RoIAlign outputs 14x14 resolution features, that are fed into 8 consequent convolutional layers (conv3x3, 256 filters, stride 1), a single transposed convolutional layer (convtranspose2d 4x4, $K$ filters, stride 2), a bilinear interpolation layer that upsamples the masks to 56x56 resolution. The output mask tensor has a size of $K$x56x56 ($K$ is a number of facial landmarks).

We emphasize that the keypoint head only slightly increases the number of calculations compared to overall feature extraction during detection (see Table \ref{tab:object_detection_gflops} and Section \ref{sec:inference_time}). If there are few faces on the image the keypoint head can be used almost for free while providing precise landmark localization.

\renewcommand{\arraystretch}{1.5}
\begin{table}[h]
	\setlength{\tabcolsep}{25.0pt}
	\scriptsize
	\centering
	\begin{tabular}{l|c}
		\hline \noalign{\smallskip}
		Feature extractor & GFLOPs\\
		\hline
		
		\hline
	  	ResNet-$50$+FPN  & $58$\\
	  	\hline
		ResNet-$50$+FPN+Context & $90$\\
		\hline
	    Keypoint head (1 proposal) & $1$\\
        \hline 

	    \hline
	\end{tabular}
	\caption{GFlops of feature extractors for an input image of 640x640. For a few faces in the image landmark localization adds small overhead to overall face detection.}
	\label{tab:object_detection_gflops}
\end{table}


\begin{figure*}[t]
\centering
    \begin{subfigure}{0.33\textwidth}
        \label{fig:ve}
        \includegraphics[width=\textwidth, trim={0 0 2cm 2cm}]{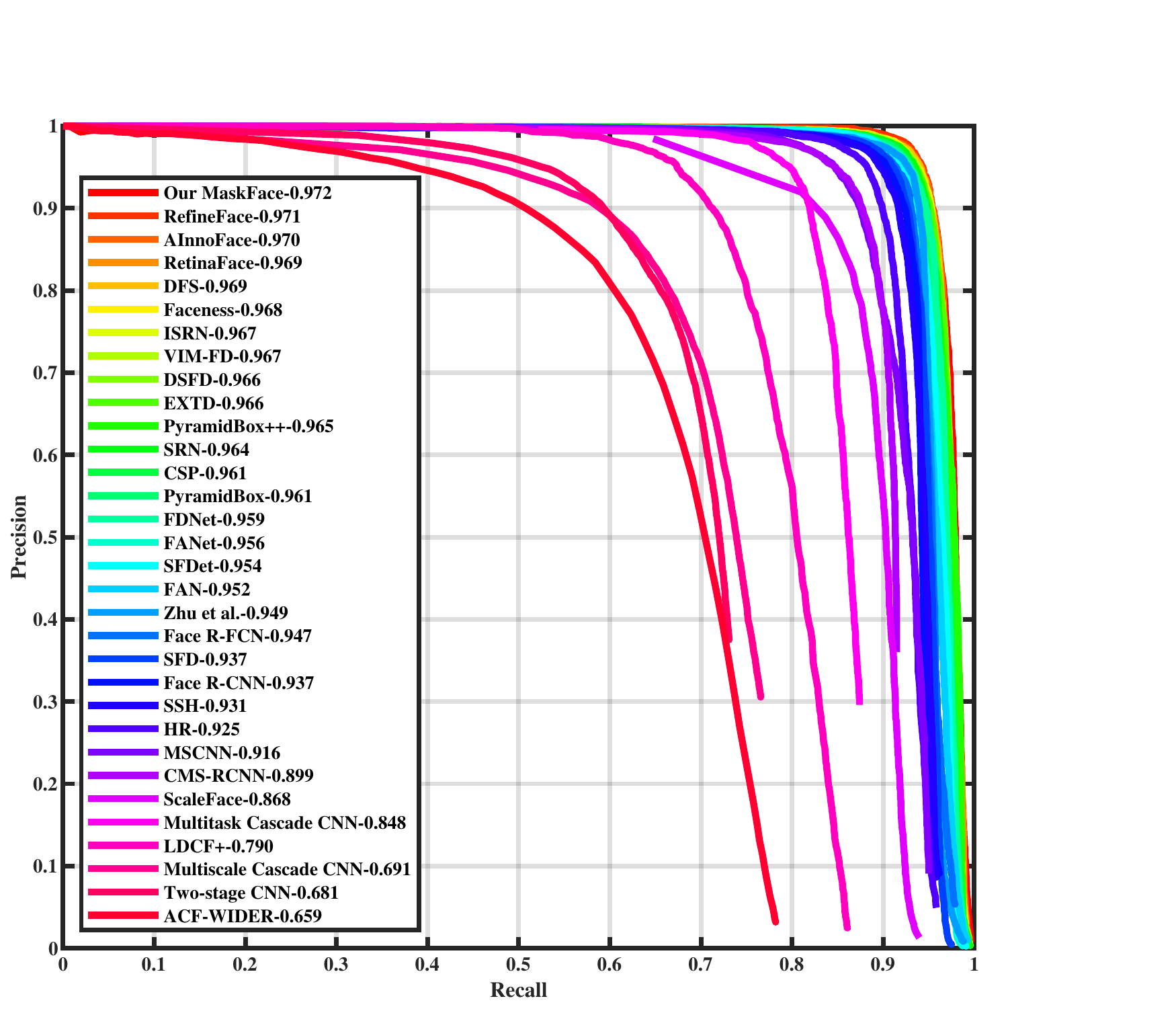}
        \centering
        \subcaption*{Val: \textit{easy}}
    \end{subfigure}%
    \hfill
    \begin{subfigure}{0.33\textwidth}
        \label{fig:vm}
        \includegraphics[width=\textwidth, trim={0 0 2cm 2cm}]{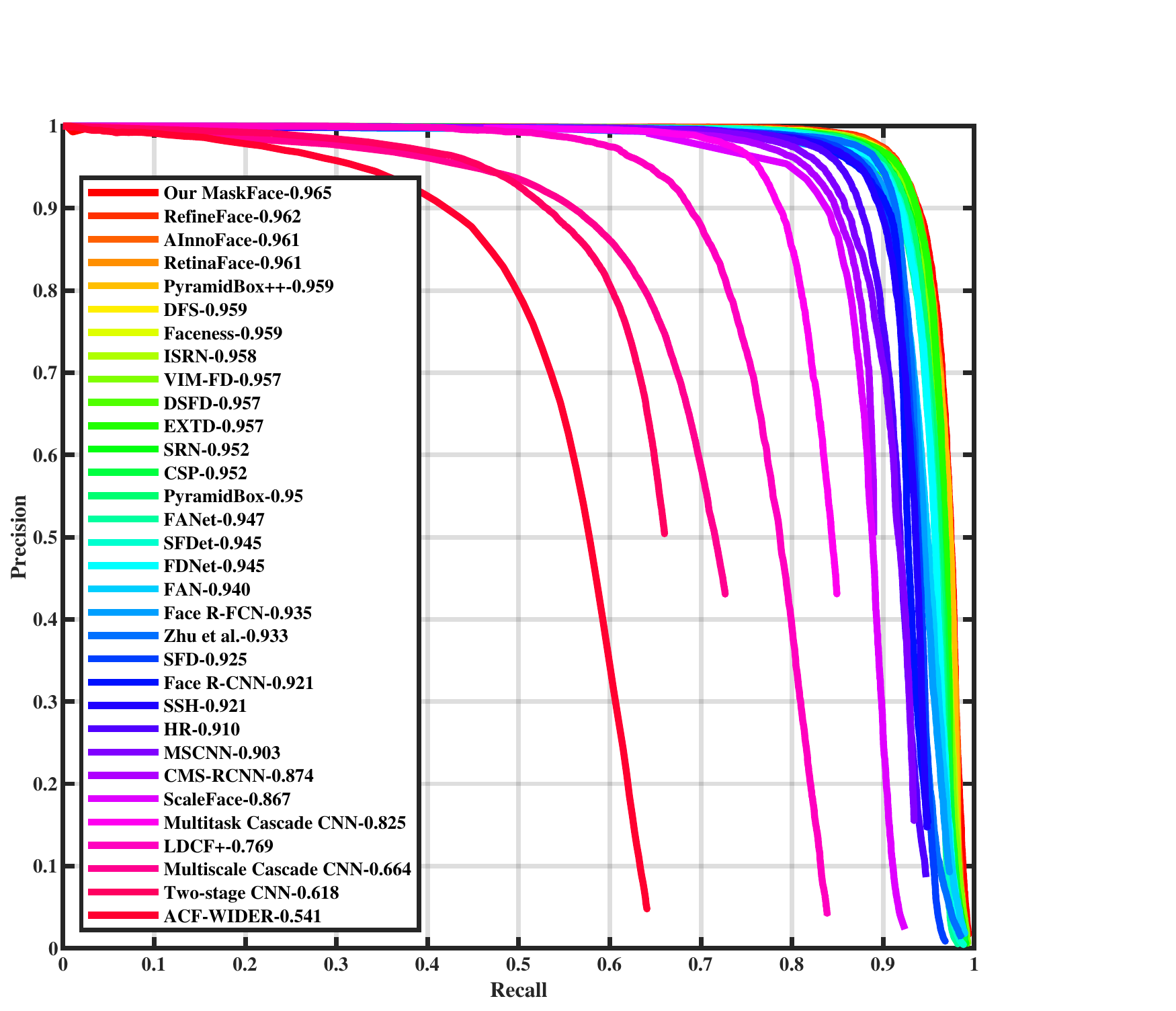}
        \subcaption*{Val: \textit{medium}}
        \centering
    \end{subfigure}%
    \hfill
    \begin{subfigure}{0.33\textwidth}
        \label{fig:vh}
        \includegraphics[width=\textwidth, trim={0 0 2cm 2cm}]{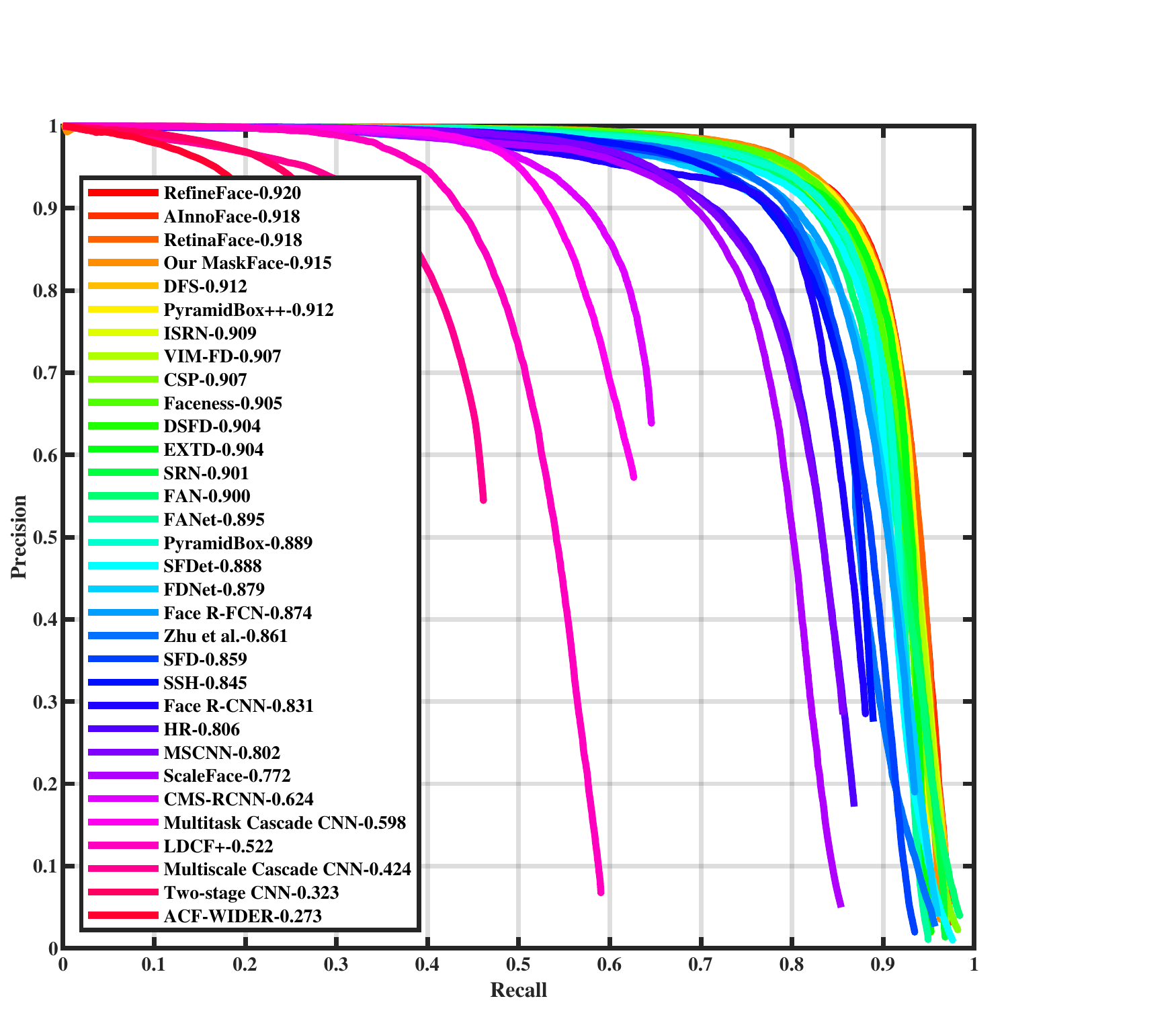}
        \subcaption*{Val: \textit{hard}}
        \centering
    \end{subfigure}
    \begin{subfigure}{0.33\textwidth}
        \label{fig:te}
        \includegraphics[width=\textwidth, trim={0 0 2cm 0cm}]{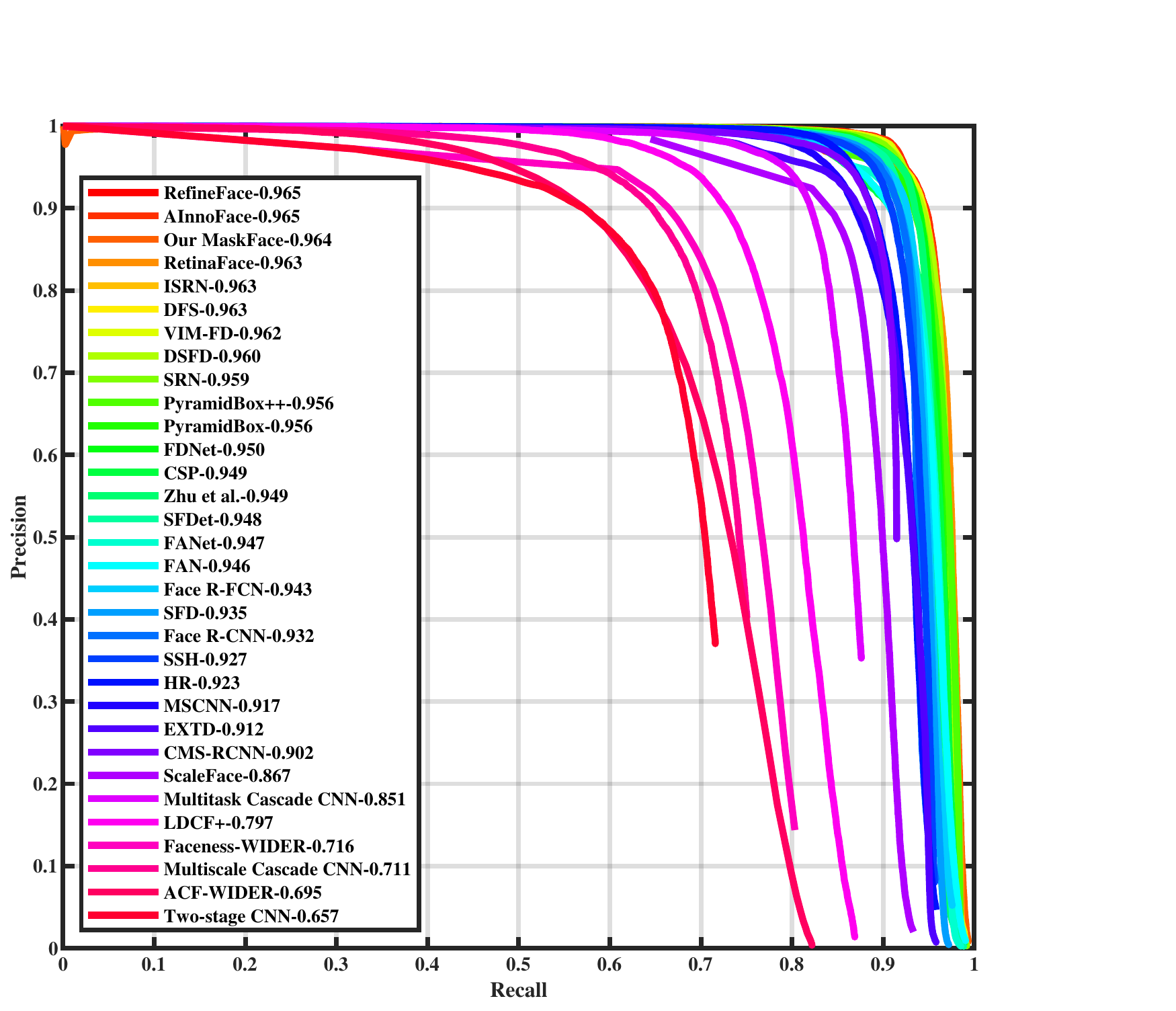}
        \subcaption*{Test: \textit{easy}}
        \centering
    \end{subfigure}%
    \hfill
    \begin{subfigure}{0.33\textwidth}
        \label{fig:tm}
        \includegraphics[width=\textwidth, trim={0 0 2cm 0cm}]{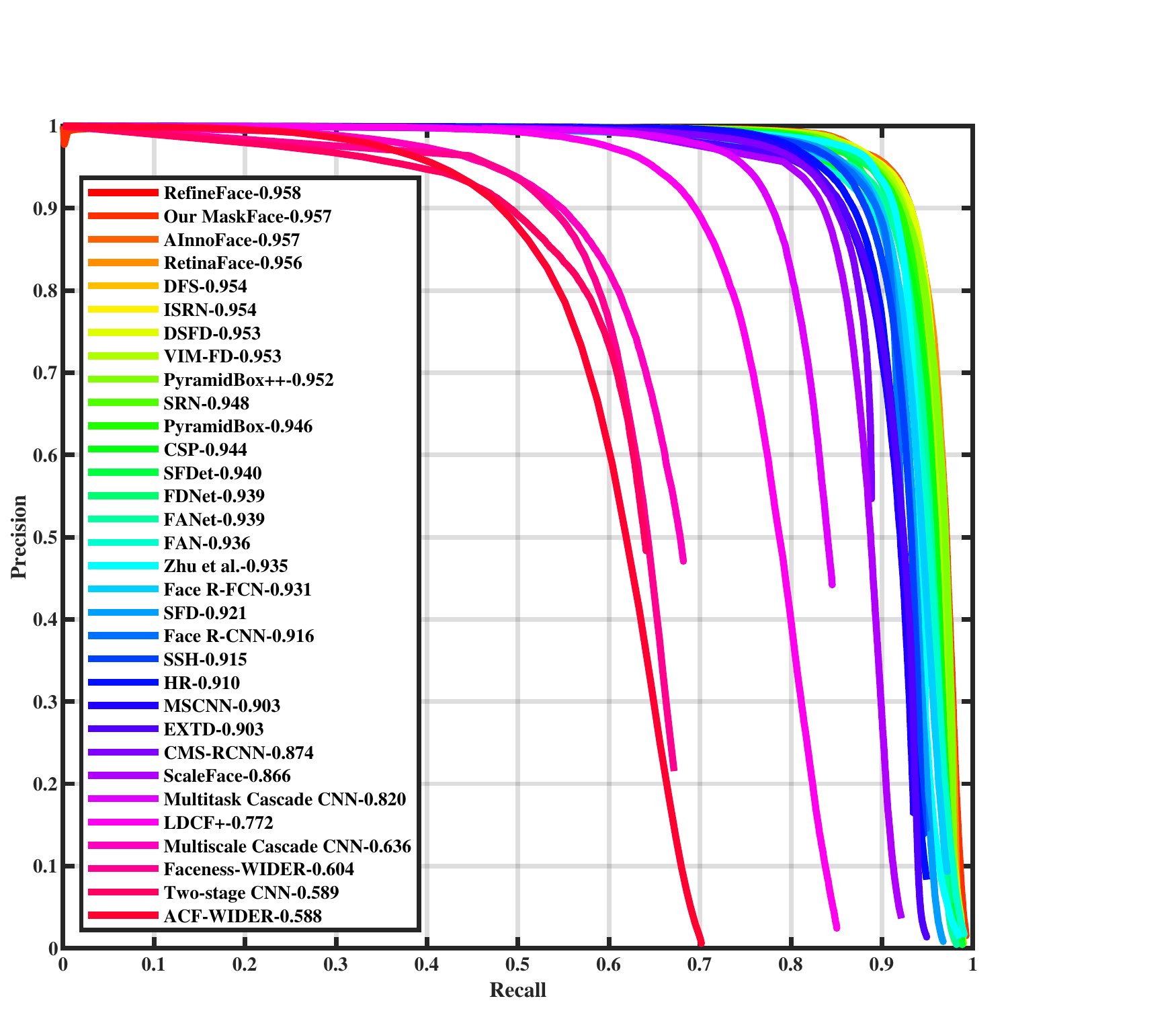}
        \subcaption*{Test: \textit{medium}}
        \centering
    \end{subfigure}%
    \hfill
    \begin{subfigure}{0.33\textwidth}
        \label{fig:th}
        \includegraphics[width=\textwidth, trim={0 0 2cm 0cm}]{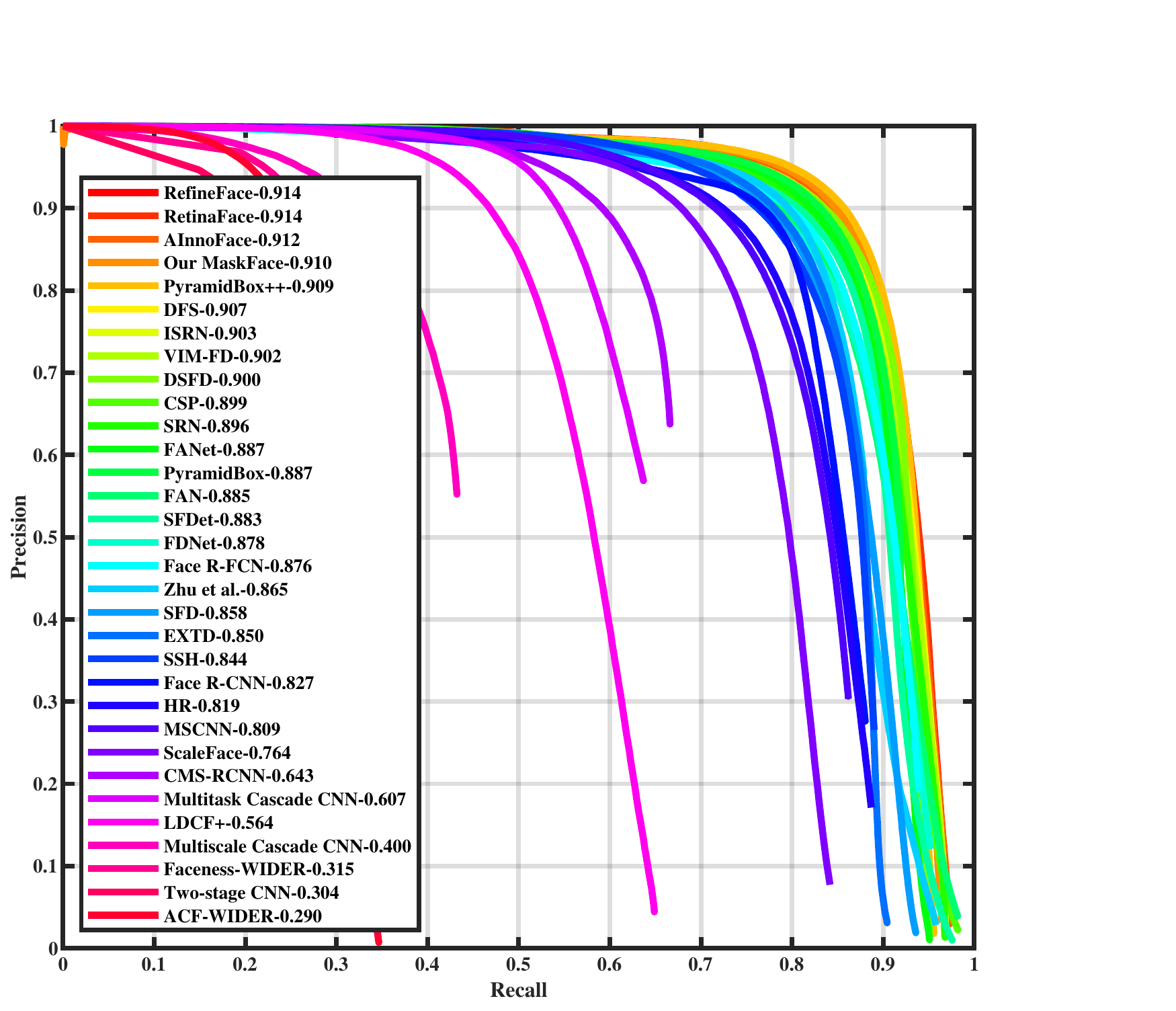}
        \centering
        \subcaption*{Test: \textit{hard}}
    \end{subfigure}
\caption{Precision-recall curves on the WIDER FACE \textit{validation} and \textit{test} subsets. The higher the better.}
\label{fig:wider-face}
\end{figure*}

\subsection{Training}

\noindent
{\bf Data augmentation.} Training on the WIDER FACE dataset. Images are resized with a randomly chosen scale factor between 0.5 and 2.5. After that we filter image annotations: we remove small and large bounding boxes with areas out of [0.4*$area_{small}$, 2.5*$area_{large}$] range, where $area_{small}$ -- an area of the smallest anchor equal to 16x16, $area_{large}$ -- an area of the largest anchor equal to 406x406. This filtering is necessary because we use the low-quality matching strategy and all ground truth boxes including ones with a small anchors' overlap are matched. Such loose matching for very small and large ground truth boxes will be harmful for training.

After that we crop patches of the 640x640 size. With probabilities of 0.5 we use either 1) random cropping or 2) random cropping around a randomly chosen bounding box. If there are no bounding boxes on the image, we perform common random cropping. Enforcing about one half of the cropped patches to contain at least one bounding box helps to enrich training batches with positive samples and increase accuracy. We apply random horizontal flipping and color distortions to the final images.

Training on the AFLW and 300W datasets. We randomly resize images so that a size of bounding boxes is in the range from 150 to 450 pixels. After that we randomly crop patches of the 480x480 size. We augment the data by $\pm 30$ degrees in-plane rotation, randomly flipping and color distortions.

\noindent
{\bf Training details.} We train MaskFace using the SGD optimizer with the momentum of 0.9 and the weight decay of 0.0001. A batch size is equal either to 8 or 16 depending on the model size. If not mentioned all backbones are pretrained on the ImageNet-1k dataset. First two convolutional layers and all batch normalization layers are frozen. We use group normalization in the feature pyramid top-down pathway, context modules and keypoint head. A number of groups is 32. We start training using warmup strategy: at first 5k iteration a learning rate grows linearly from 0.0001 to 0.01. After that to update the learning rate we apply a step decay schedule. When accuracy on validation set stops growing, we multiply the learning rate by a factor of 0.1. The minimum learning rate is 0.0001.

\section{Experiments}

\begin{figure*}[t]
\centering
    \begin{subfigure}{0.33\textwidth}
        \includegraphics[width=\textwidth,trim={0.5cm 1cm 2cm 2cm}]{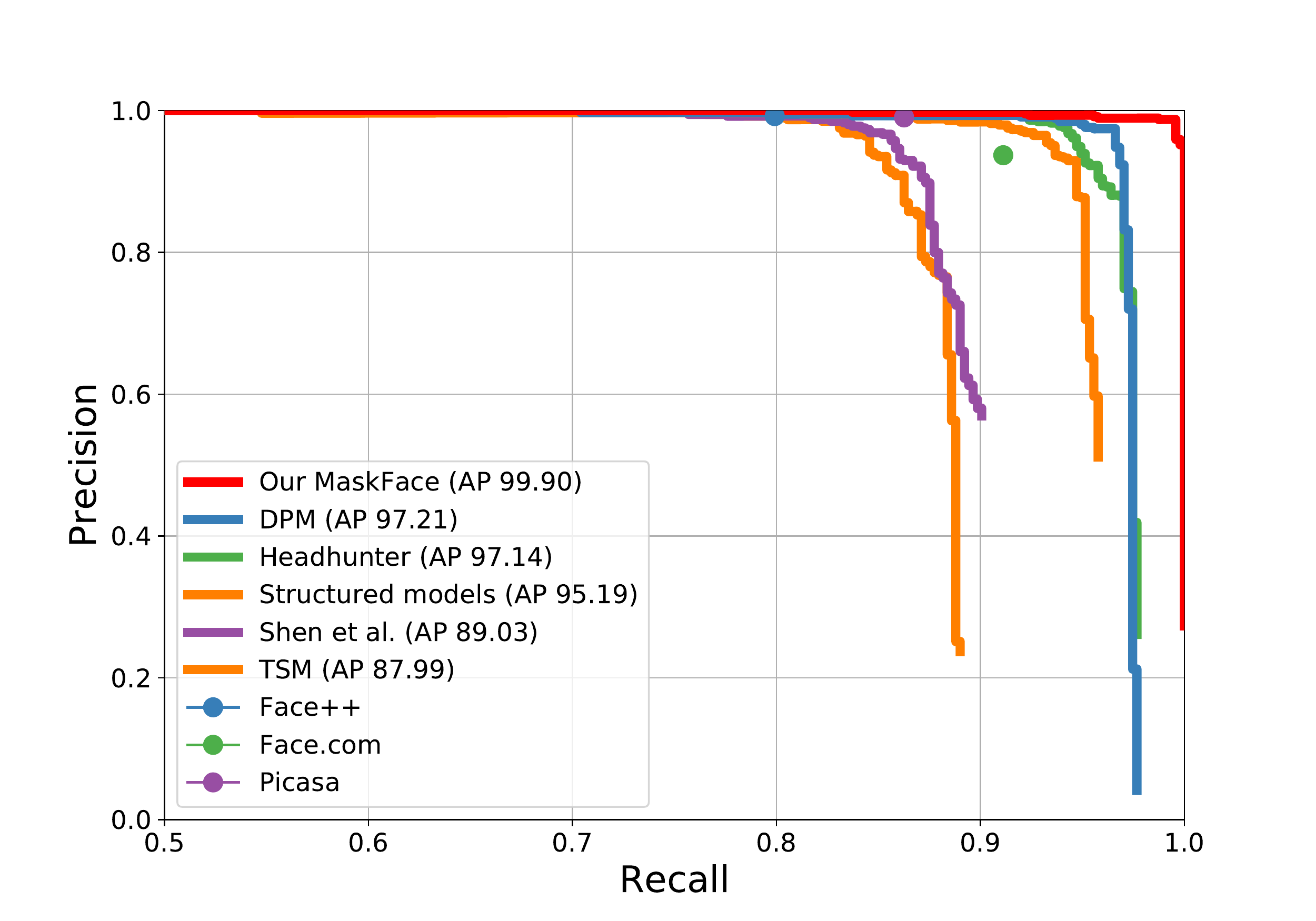}
        \centering
        \subcaption*{AFW}
        \label{fig:AFW}
    \end{subfigure}%
    \hfill
    \begin{subfigure}{0.33\textwidth}
        \includegraphics[width=\textwidth,trim={0.5cm 1cm 2cm 2cm}]{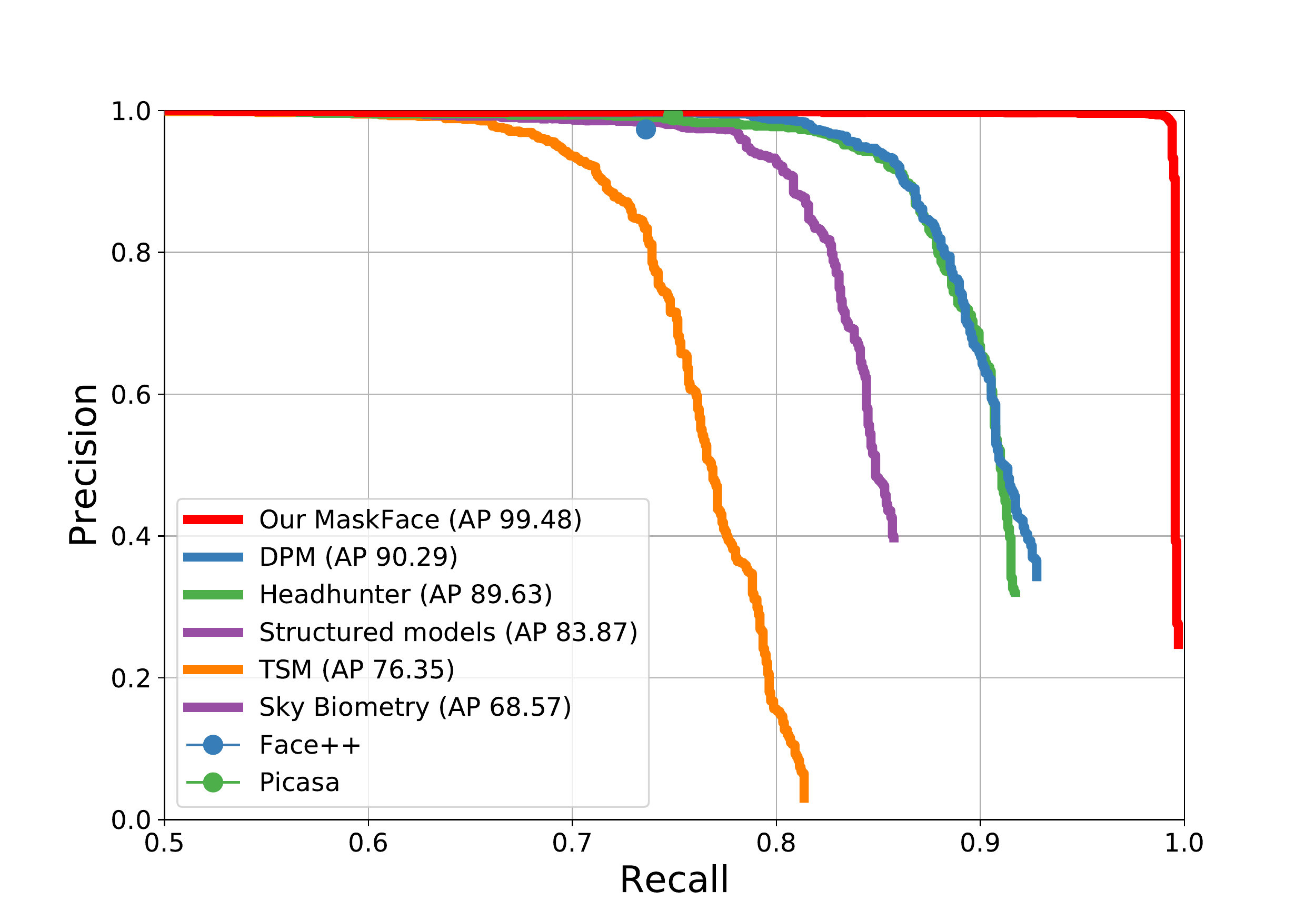}
        \centering
        \subcaption*{PASCAL}
        \label{fig:FDDB}
    \end{subfigure}
    \hfill
    \begin{subfigure}{0.33\textwidth}
        \includegraphics[width=\textwidth,trim={1cm 1.5cm 2.5cm 2.5cm}]{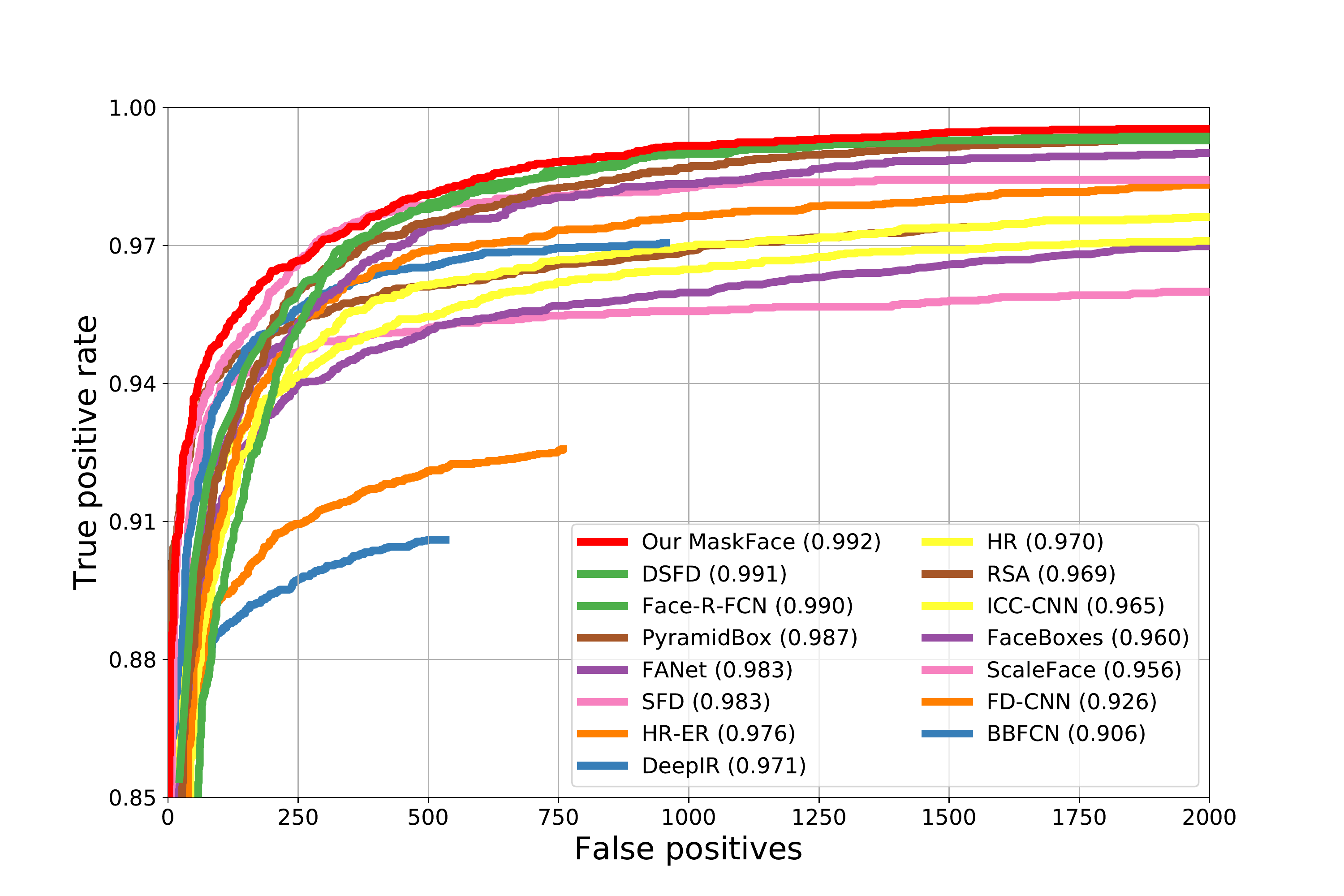}
        \centering
        \subcaption*{FDDB}
        \label{fig:FDDB}
    \end{subfigure}
\caption{Precision-recall curves for the AFW, PASCAL and FDDB datasets. The higher the better.}
\label{fig:face-datasets}
\end{figure*}

\subsection{Datasets}

\noindent
{\bf WIDER FACE.} The WIDER FACE dataset \cite{yang2016wider} contains 32203 images and 393703 labeled face bounding boxes with a high degree of variability in scale, pose and occlusion. Each subset contains three levels of difficulty: \textit{easy}, \textit{medium} and \textit{hard}. Face bounding box annotations are provided for \textit{train} and \textit{validation} subsets with 12880 and 3226 images respectively. In \cite{deng2019retinaface} the authors release annotations of 5 facial landmarks (left and right eyes, nose, left and right mouth corners) for the WIDER FACE \textit{train} subset. In total they provide annotations for 84600 faces. At the time of paper submission, the authors did not provide annotations for the \textit{validation} subset.

\noindent
{\bf AFW.} The AFW dataset \cite{zhu2012face} has 205 images with 473 faces. The images in the dataset contains cluttered backgrounds with large variations in both face viewpoint and appearance.

\noindent
{\bf PASCAL face.} The PASCAL dataset \cite{yan2014face} is collected from the test set of PASCAL person layout dataset, consisting of 1335 faces with large face appearance and pose variations from 851 images.

\noindent
{\bf FDDB.} The FDDB dataset \cite{jain2010fddb} has 5171 faces in 2845 images taken from news articles on Yahoo websites. Faces in FDDB are represented by ellipses and not all of them are marked. Therefore, for evaluation we use the annotation with additional labeled faces from the SFD paper \cite{zhang2017s3fd}.

\noindent
{\bf AFLW.} The AFLW dataset \cite{koestinger2011annotated} contains 21997 real-world images with 25993 annotated faces. Collected images have a large variety in face appearance (pose, expression, ethnicity, age, gender) and environmental conditions. Each face annotation includes a bounding box and up to 21 visible landmarks. In \cite{zhu2016unconstrained} the authors revised the AFLW annotation and provided labeling of all 19 landmarks (regardless its visibility) and a visibility flag. In our experiments we use the revised annotation.

\noindent
{\bf 300W.} The 300W dataset \cite{sagonas2013300} is a combination of HELEN \cite{le2012interactive}, LFPW \cite{belhumeur2013localizing}, AFW \cite{zhu2012face}, XM2VTS and IBUG datasets, where each face has 68 landmarks. We follow the same evaluation protocol as described in HRNet \cite{sun2019high}. We use 3148 training images, which contains the training subsets of HELEN and LFPW and the \textit{full} set of AFW. We evaluate the performance using two protocols, \textit{full} set and
\textit{test} set. The \textit{full} set contains 689 images and is further divided into a \textit{common} subset (554 images) from HELEN and LFPW, and a \textit{challenging} subset (135 images) from IBUG. The official \textit{test} set, used for competition, contains 600 images (300 indoor and 300 outdoor images).

\subsection{Evaluation}

{\bf Face detection.} We use common test time image augmentations: horizontal flipping and the image pyramid. The multi-scale is essential for detecting tiny faces. We apply Soft-NMS \cite{bodla2017soft} to bounding box predictions for each augmented image. After that all bounding box predictions from the augmented images are joined and filtered by the box voting \cite{gidaris2015object}. For evaluation on WIDER FACE and FDDB we use the officially provided toolboxes \cite{yang2016wider,jain2010fddb}. For evaluation on AFW and PASCAL we use the toolbox from \cite{mathias2014face}.

\noindent
{\bf Landmark localization.} We calculate normalized mean error (NME) metrics using the face bounding box as normalization for the AFLW dataset and the inter-ocular distance as normalization for 300W. When we make comparison with multi-task approaches an input image is resized to make the minimum image size equal to 640. When we compare our method with single-task approaches images are cropped and resized so that ground truth faces has a size of 256x256 following to HRNet \cite{sun2019high}. We emphasize that for the landmark evaluation we do not apply any test time image augmentations as well as any kind of predictions’ averaging. MaskFace outputs several proposals per a ground truth face, but we choose only one the most confident proposal for landmark predictions.

\subsection{Main results}
\noindent
{\bf Face detection.} All results are provided for the ResNeXt-152 backbone pretrained on the ImageNet-5k and COCO datasets \cite{he2017mask,Detectron2018}. The model is trained on the WIDER FACE with 5 facial keypoints and evaluated on the AFW, PASCAL face, FDDB datasets. In Figure \ref{fig:wider-face} we show precision-recall curves of the proposed approach for the WIDER FACE \textit{validation} and \textit{test} subsets. In Figure \ref{fig:face-datasets} we show precision-recall for the AFW, PASCAL and FDDB datasets. We could not find precision-recall curves for recent state-of-the-art detectors on AFW and PASCAL, therefore in Table \ref{tab:object_detection_accuracy} we additionally show AP metrics collected from papers. Our experiments demonstrate that MaskFace achieves state-of-the-art results. 

\renewcommand{\arraystretch}{1.5}
    \begin{table}[h]
 	\setlength{\tabcolsep}{4.5pt}
 	\scriptsize
 	\centering
 	\begin{tabular}{l|c}
 		\hline \noalign{\smallskip}
 		Method & AP \\
 		\hline
	
 		\hline
 	  	BB-FCN \cite{liu2019facial} & $97.46$\\
 	  	\hline
 	  	FaceBoxes \cite{zhang2017faceboxes} & $98.91$\\
 	  	\hline
 	  	SFD\cite{zhang2017s3fd} & $99.85$\\
 	  	\hline
 	  	SRN \cite{chi2018selective} & $99.87$\\
 	  	\hline
 	  	{\bf Our MaskFace} & $\mathbf{99.90}$\\
 	  	\hline

 	    \hline

 	\end{tabular}
 	\quad
 	\begin{tabular}{l|c}
 	    \hline \noalign{\smallskip}
 		Method & AP \\
 		\hline
	
 		\hline
 	    FaceBoxes \cite{zhang2017faceboxes} & $96.30$\\
 	  	\hline
 	  	SFD \cite{zhang2017s3fd} & $98.49$\\
 	  	\hline
 	  	SRN \cite{chi2018selective} & $99.09$\\
 	  	\hline
 	  	{\bf Our MaskFace} & $\mathbf{99.48}$\\
 	  	\hline

 	    \hline
 	\end{tabular}

 	\caption{AP metrics on the AFW (left) and PASCAL (right) datasets. The higher the better.}
  	\label{tab:object_detection_accuracy}
 	\end{table}

\begin{figure}[h]
    \includegraphics[width=0.52\textwidth,trim={1cm 2cm 0cm 2.5cm}]{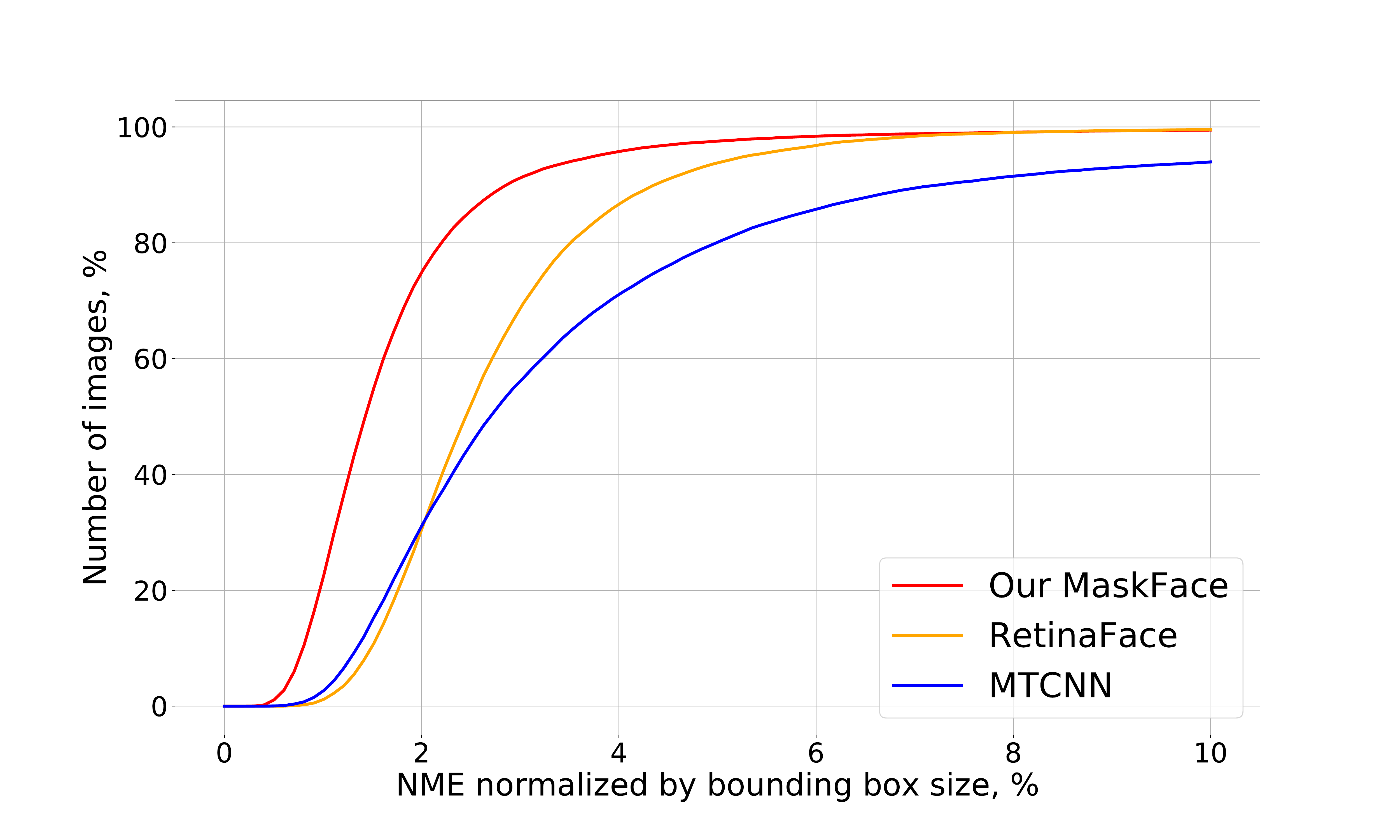}
    \centering
    \caption{Comparison with multi-task methods. Cumulative distribution for NME of 5 facial landmarks on the \textit{full} AFLW dataset. The higher the better.}
    \label{fig:Full_AFLW_CED}
\end{figure}

\begin{figure*}[h]
\centering
    \begin{subfigure}{0.33\textwidth}
        \includegraphics[width=\textwidth, trim={20cm 16cm 20cm 5cm}, clip]{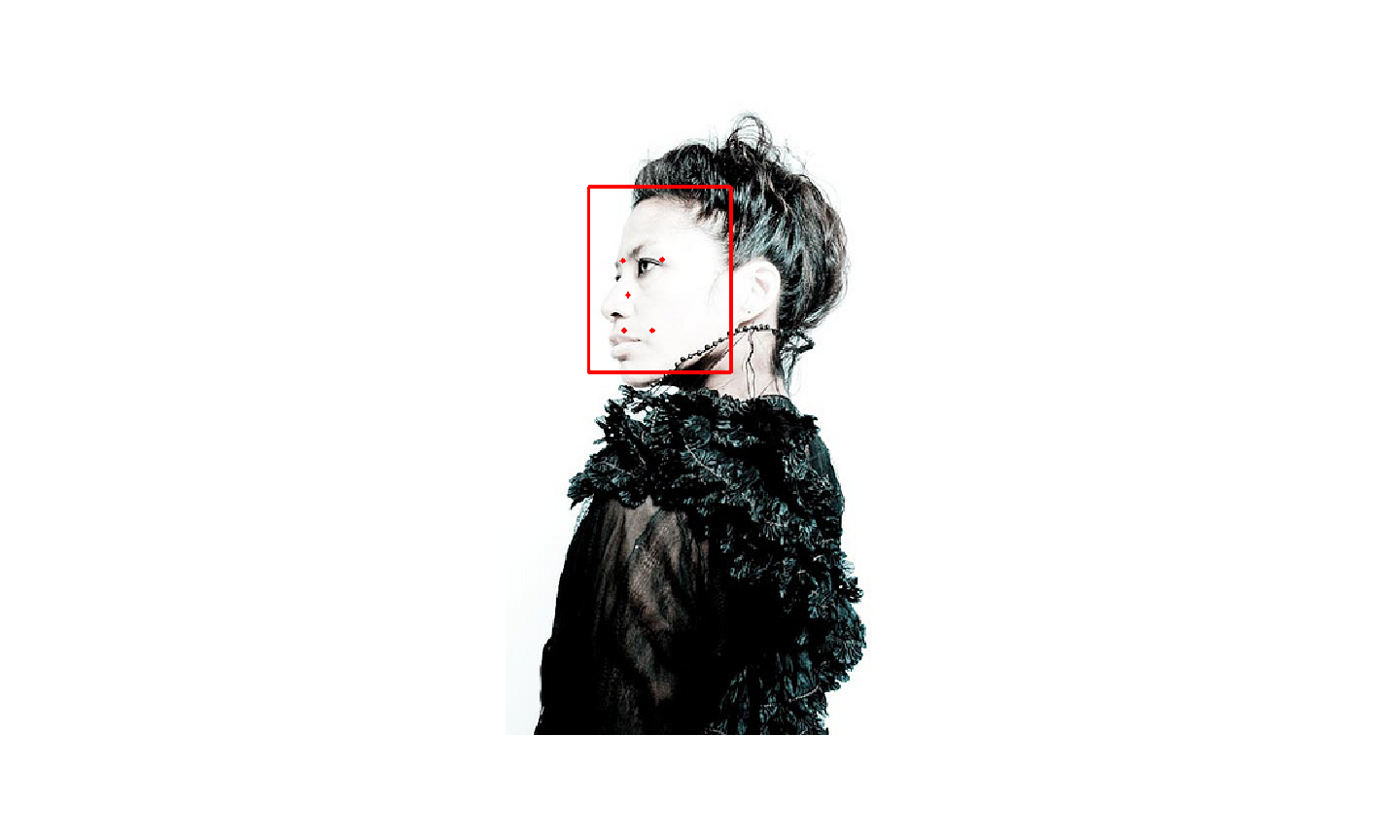}
        \centering
        \caption*{MTCNN}
        \label{mtcnn_woman}
    \end{subfigure}%
    \hfill
    \begin{subfigure}{0.33\textwidth}
        \includegraphics[width=\textwidth, trim={20cm 16cm 20cm 5cm}, clip]{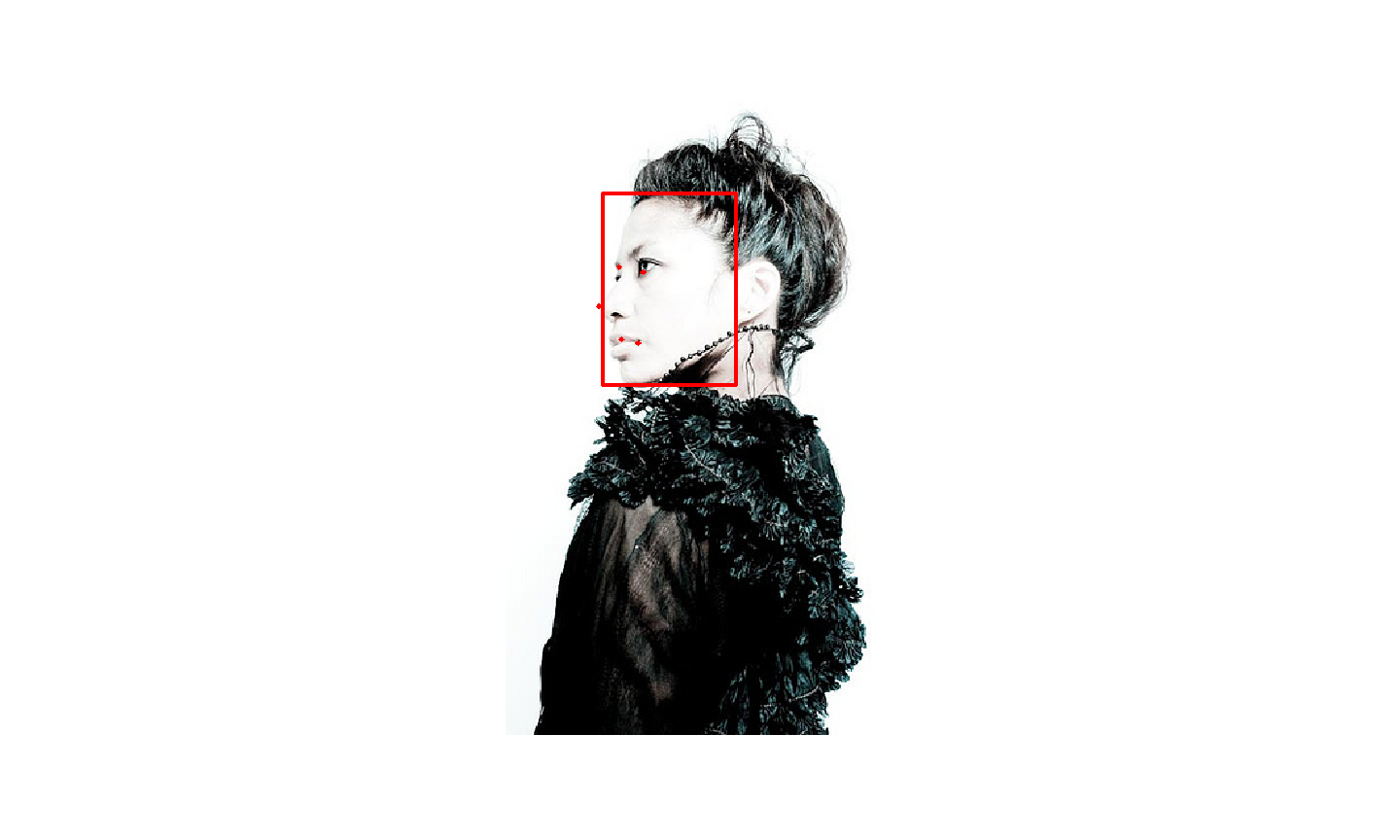}
        \centering
        \caption*{RetinaFace}
        \label{retinaface_woman}
    \end{subfigure}%
    \begin{subfigure}{0.33\textwidth}
        \includegraphics[width=\textwidth, trim={20cm 16cm 20cm 5cm}, clip]{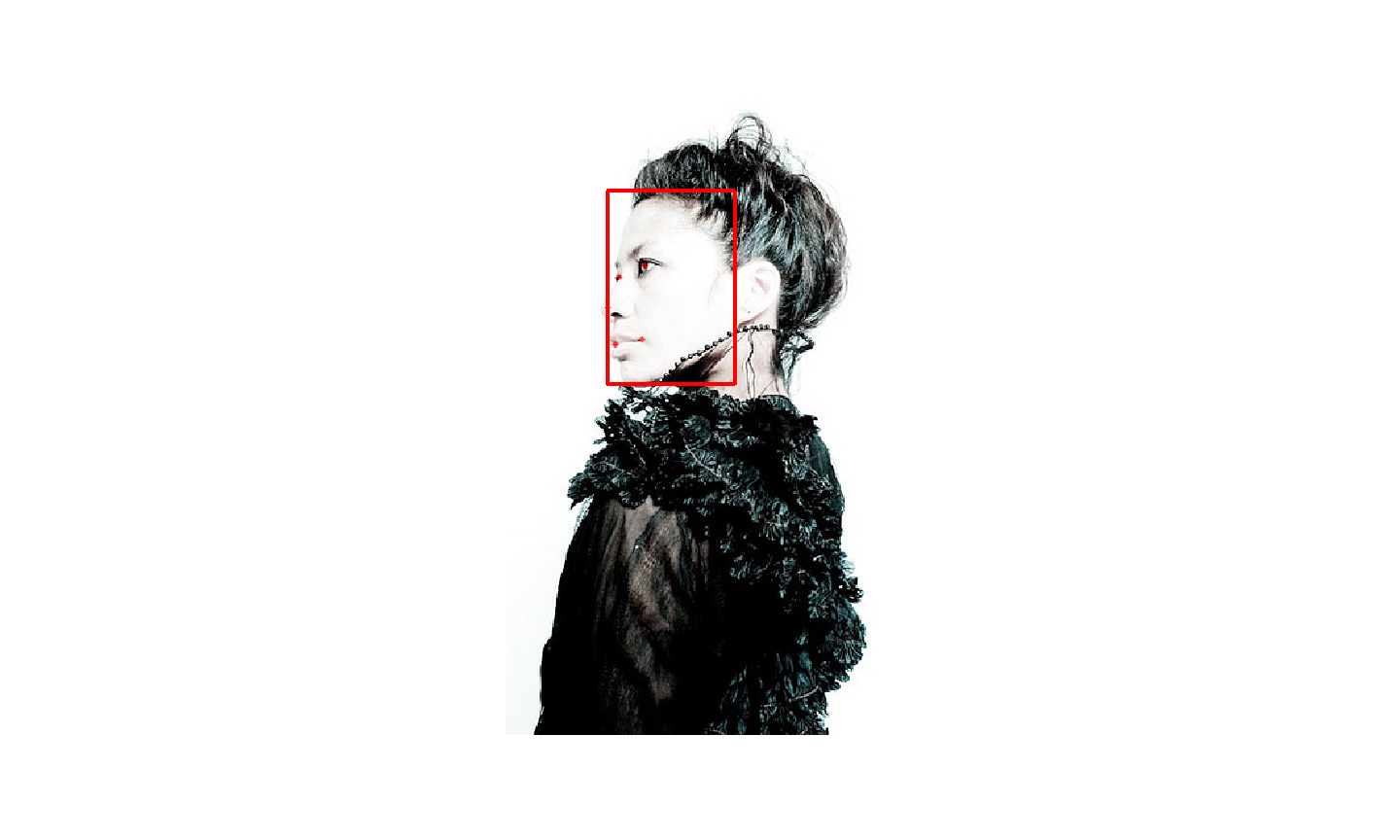}
        \centering
        \caption*{MaskFace}
        \label{maskface_woman}
    \end{subfigure}
    \caption{Illustration of differences in localization of 5 facial landmarks for MTCNN, RetinaFace and MaskFace on AFLW.}
\label{fig:5_landmark_comparison}
\end{figure*}

\noindent
{\bf Landmark localization.} First, we compare our approach with the state-of-the-art multi-task face detector RetinaFace \cite{deng2019retinaface} and popular MTCNN \cite{zhang2016joint}. To make fair comparison with RetinaFace we train MaskFace using the same ResNet-50 backbone on WIDER FACE. RetinaFace implementation and trained weights are taken from \cite{deng2019retinaface}. To match predicted boxes with ground truth, we use an IoU threshold of 0.3. In Figure \ref{fig:Full_AFLW_CED} we plot cumulative error distribution (CED) curves for NME of 5 facial landmarks on the full AFLW dataset (21997 images). For qualitative comparison see Figure \ref{fig:5_landmark_comparison}. The CED curves characterize distributions of errors for each method, that can be also shown by the histogram of errors. If any of the methods cannot detect some faces, then it will affect only the upper part of the corresponding CED curve (the upper part will be cut off). Because usually if a method cannot detect some faces then such faces have a high value of NME. Figure \ref{fig:Full_AFLW_CED} shows that MaskFace outperforms previous multi-task approaches by a large margin and improves the baseline on the AFLW dataset for sparse landmarks.

\renewcommand{\arraystretch}{1.5}
\begin{table}[h]
\setlength{\tabcolsep}{9.0 pt}
\scriptsize
\centering
\begin{tabular}{l|l|c|c }
\hline \noalign{\smallskip}
 & Backbone &  Full &  Frontal \\
\hline

\hline
TSR \cite{lv2017deep}&VGG-S &$2.17$ & - \\
CPM + SBR \cite{dong2018supervision}& CPM &$2.14$&-\\
SAN \cite{dong2018style}& ResNet-$152$ &$1.91$ & $1.85$\\
DSRN \cite{miao2018direct}& - &$1.86$ & -\\
LAB (w/o B) \cite{wu2018look}& Hourglass & $1.85$ & $1.62$\\
HRNet \cite{sun2019high}& HRNetV2-W$18$ & $1.57$ & $1.46$ \\
{\bf Our MaskFace} & ResNet-50 + Context & $\mathbf{1.54}$ & $\mathbf{1.31}$\\
\hline

\hline
\multicolumn{3}{l}{
\emph {Model trained with \texttt{extra} info.}}\\
\hline
DCFE (w/ $3$D)~\cite{valle2018deeply}& - &$2.17$ & - \\
PDB (w/ DA)~\cite{feng2018wing}& ResNet-$50$&$1.47$ & -\\
LAB (w/ B)~\cite{wu2018look}& Hourglass &$1.25$ & $1.14$\\
\hline
\end{tabular}
\caption{Detection results (NME) for 19 facial landmarks on the AFLW \textit{test} subset. The lower the better.}
\label{table:comparison_aflw_testset}
\end{table}

\renewcommand{\arraystretch}{1.3}
\begin{table}
\scriptsize
\setlength{\tabcolsep}{4pt}
\centering
\begin{tabular}{l|l|cccc}
\hline\noalign{\smallskip}
Method & Backbone & Common & Challenging & Full & Test\\
\hline

\hline
RCN \cite{honari2016recombinator} &-& $4.67$ & $8.44$ & $5.41$ & -\\
DSRN \cite{miao2018direct} &-& $4.12$ & $9.68$ & $5.21$ & -\\
CFAN \cite{zhang2014coarse}& - & - & - & - &$5.78$\\
SDM \cite{xiong2013supervised}& - & - & - & - &$5.83$\\
CFSS \cite{zhu2015face}& - & - & - & - &$5.74$\\
PCD-CNN \cite{kumar2018disentangling} &-& $3.67$ & $7.62$ & $4.44$ & -\\
CPM + SBR \cite{dong2018supervision} & CPM & $3.28$ & $7.58$ & $4.10$ & - \\
SAN \cite{dong2018style} &ResNet-152& $3.34$ & $6.60$ & $3.98$ & - \\
MDM \cite{trigeorgis2016mnemonic}& - & - & - & - & $4.78$\\
DAN \cite{kowalski2017deep} &-& $3.19$ & $5.24$ & $3.59$ & $4.30$ \\
{\bf Our MaskFace} & ResNet-50 & $\mathbf{2.99}$ & $\mathbf{5.71}$ & $\mathbf{3.52}$ & $\mathbf{4.21}$\\
 &  + context & & & & \\
Chen et al. \cite{chen2017adversarial}& Hourglass & - & - & - &$3.96$ \\
HRNet & HRNetV$2$-W$18$ &$2.87$ & $5.15$ & $3.32$ & $3.85$\\
\hline

\hline
\multicolumn{4}{l}{
\emph {Model trained with \texttt{extra} info.}}\\
\hline
LAB (w/ B) \cite{wu2018look}& Hourglass & $2.98$\\
DCFE (w/ $3$D) \cite{valle2018deeply}& - &$2.76$& - & - & $3.88$\\
\hline

\end{tabular}
\caption{Detection results (NME) for 68 facial landmarks on the 300W subsets. The lower the better.}
\label{table:comparison_300w_fullset}
\end{table}

\begin{figure*}[h]
\centering
    \begin{subfigure}{0.248\textwidth}
        \label{fig:ve}
        \includegraphics[width=\textwidth, trim={1.5cm 2.8cm 1.8cm 3cm},clip]{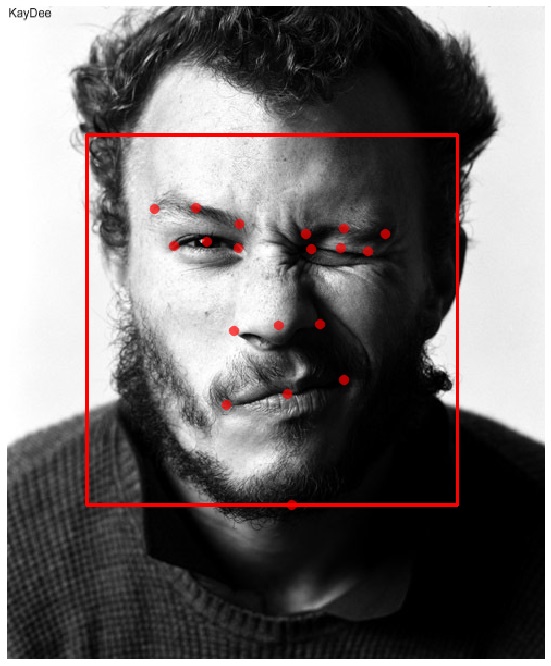}
        \centering
    \end{subfigure}%
    \hfill
    \hspace{0.0 mm}
    \begin{subfigure}{0.247\textwidth}
        \label{fig:vm}
        \includegraphics[width=\textwidth, trim={7cm 0 4cm 3cm},clip]{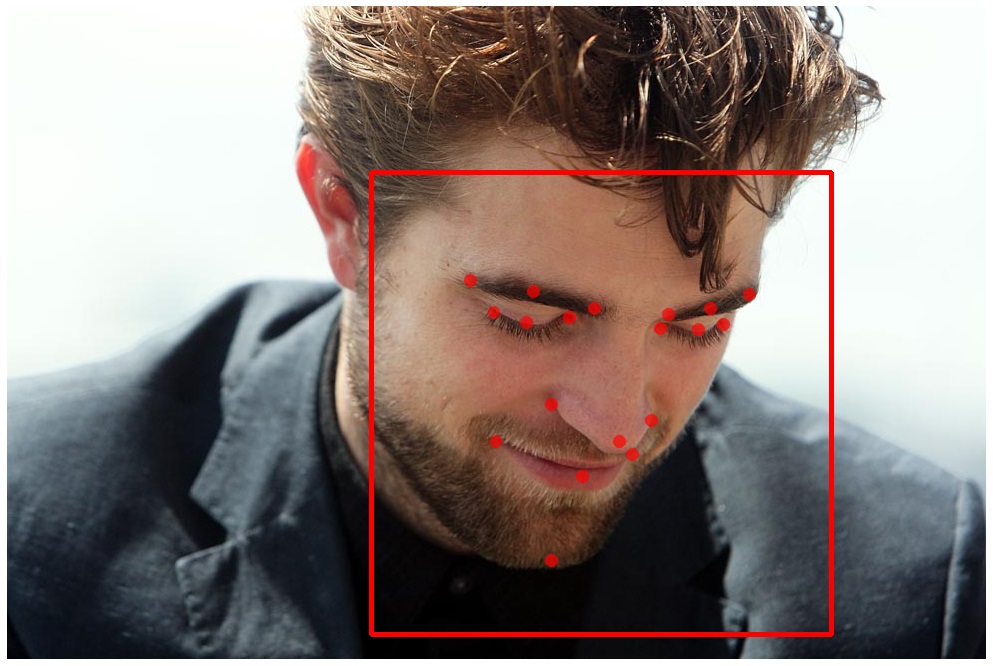}
        \centering
    \end{subfigure}%
    \hfill
    \hspace{0.0 mm}
    \begin{subfigure}{0.245\textwidth}
        \label{fig:vh}
        \includegraphics[width=\textwidth, trim={6cm 3cm 5.5cm 5cm},clip]{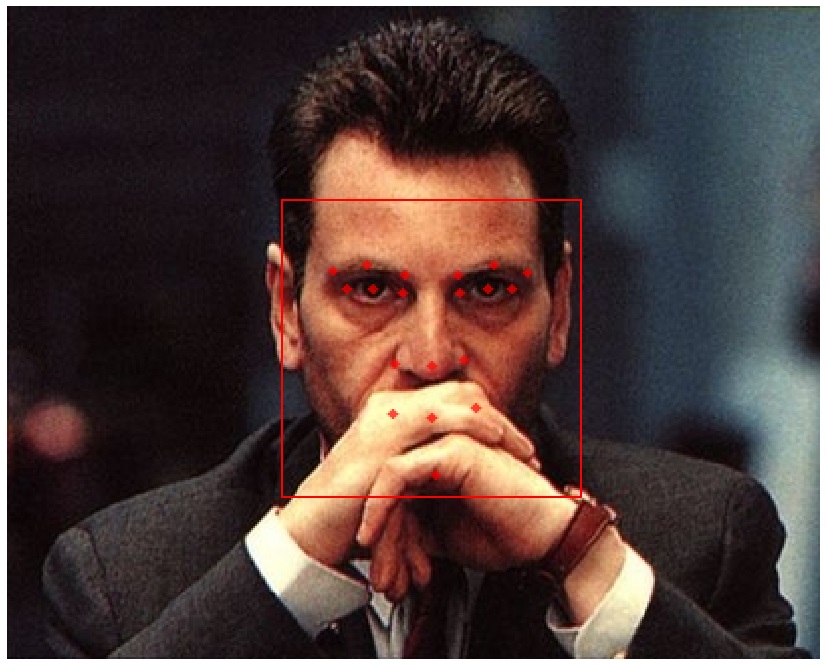}
        \centering
    \end{subfigure}%
    \hspace{0.0 mm}
    \begin{subfigure}{0.245\textwidth}
        \label{fig:vh}
        \includegraphics[width=\textwidth, trim={2cm 1cm 5cm 2cm},clip]{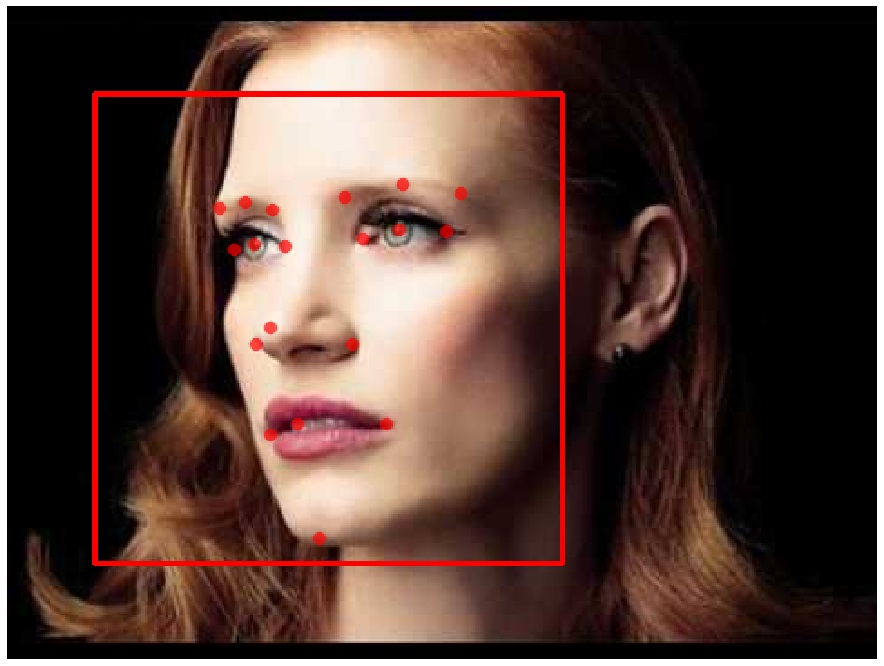}
        \centering
    \end{subfigure}
    \begin{subfigure}{0.248\textwidth}
        \label{fig:te}
        \includegraphics[width=\textwidth, trim={1.5cm 2.8cm 1.8cm 3cm},clip]{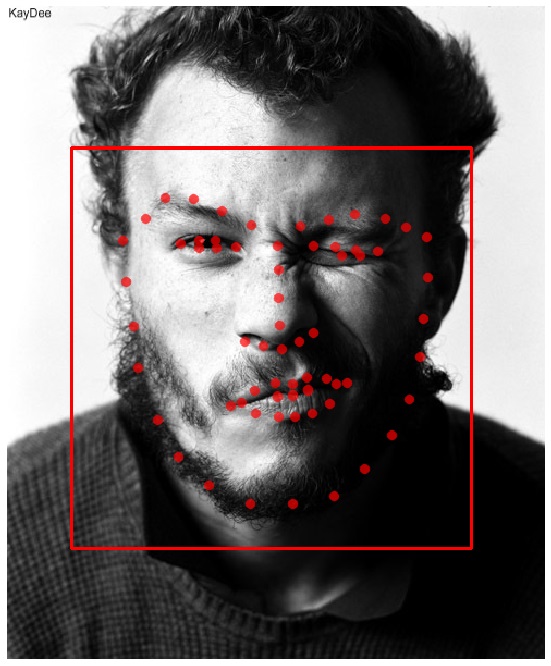}
        \centering
    \end{subfigure}%
    \hfill
    \hspace{0.0 mm}
    \begin{subfigure}{0.247\textwidth}
        \label{fig:tm}
        \includegraphics[width=\textwidth, trim={7cm 0 4cm 3cm},clip]{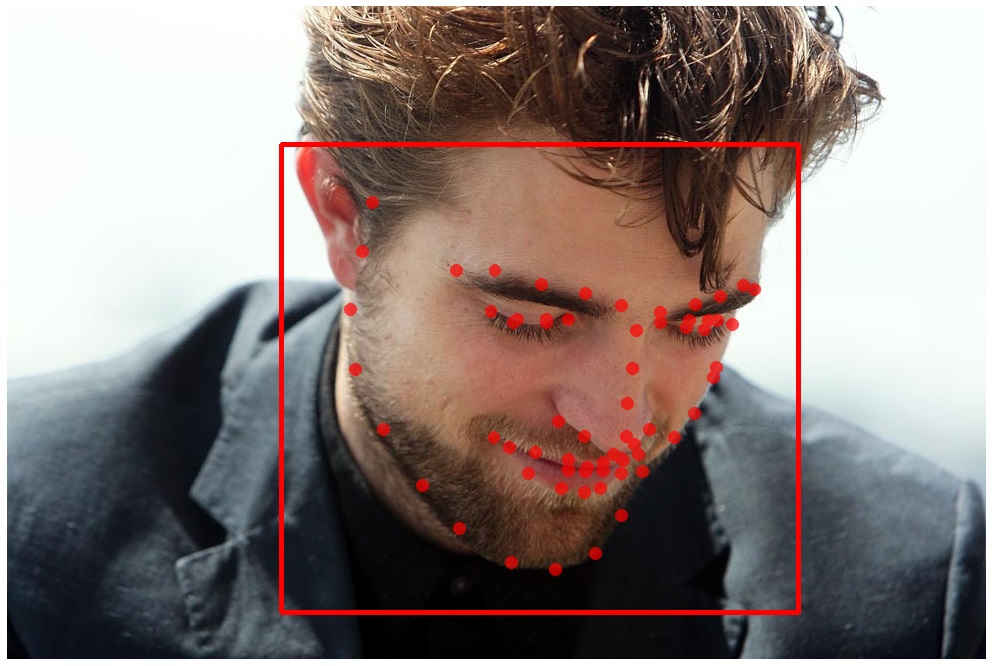}
        \centering
    \end{subfigure}%
    \hfill
    \hspace{0.0 mm}
    \begin{subfigure}{0.245\textwidth}
        \label{fig:th}
        \includegraphics[width=\textwidth, trim={6cm 3cm 5.5cm 5cm},clip]{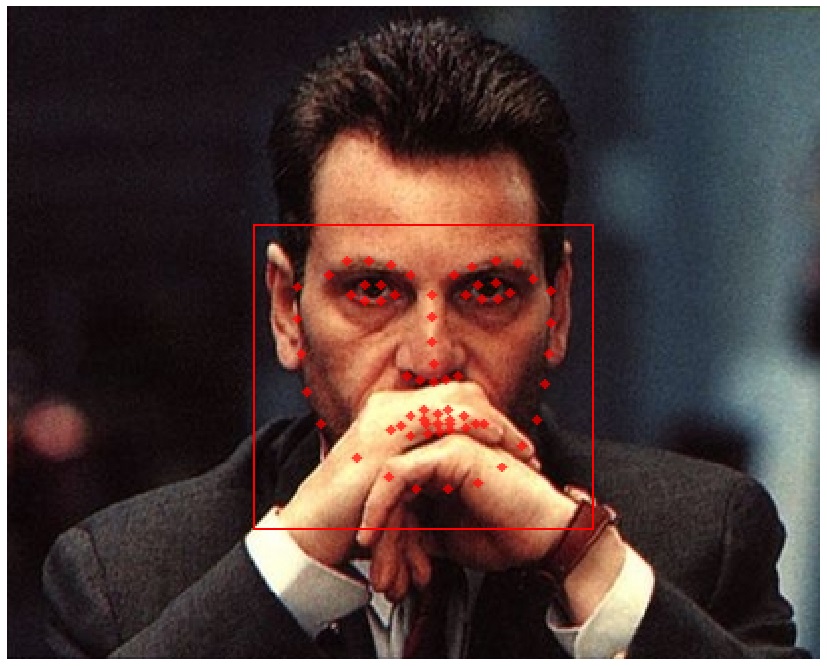}
        \centering
    \end{subfigure}%
    \hfill
    \hspace{0.0 mm}
    \begin{subfigure}{0.245\textwidth}
        \label{fig:tm}
        \includegraphics[width=\textwidth, trim={2cm 1cm 5cm 2cm},clip]{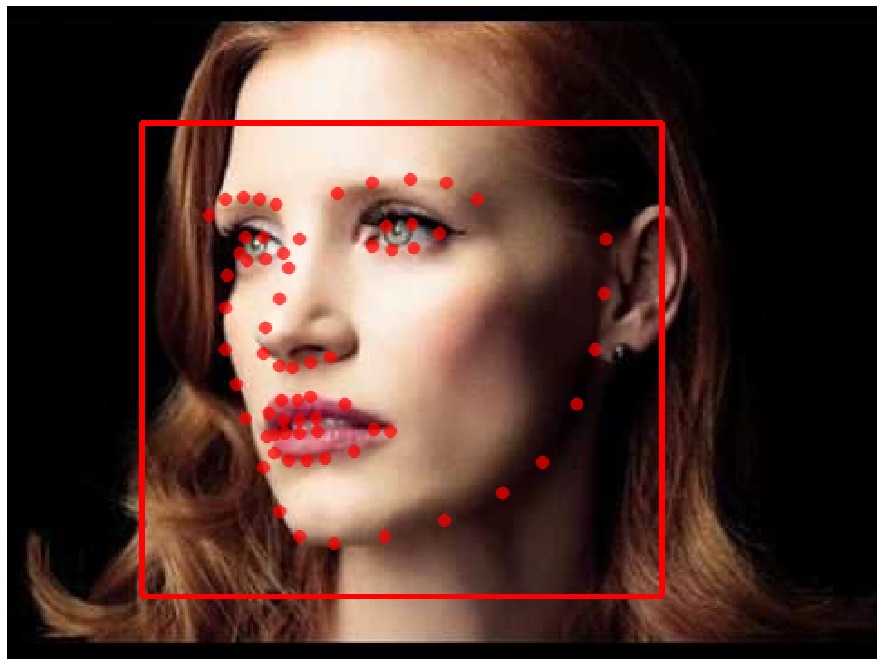}
        \centering
    \end{subfigure}
\caption{Visualization of 19 and 68 landmarks predicted by MaskFace on 300W.}
\label{fig:19_68_landmarks}
\vspace{0.55cm}
\end{figure*}

Second, we compare our method with recent state-of-the-art single-task models. The MaskFace model is trained using the ResNet-50 backbone. For training and validation we use the same AFLW and 300W subsets as in the HRNet paper \cite{sun2019high}. We pretrain the model on WIDER FACE and AFLW for evaluation on AFLW and 300W, respectively. Pretraining helps to slightly improve the accuracy. For matching predicted boxes with ground truth, we use an IoU threshold of 0.4. In Tables \ref{table:comparison_aflw_testset} and \ref{table:comparison_300w_fullset} we provide results for comparison with other methods. The obtained values of NME are given for the case when MaskFace detects \textit{all} faces. Visualization of MaskFace predictions is shown in Figure \ref{fig:19_68_landmarks}. Our approach achieves top results among methods without extra information and stronger data augmentation. Note, that LAB that uses extra boundary information (w/B) \cite{wu2018look}, PDB uses stronger data augmentation (w/DA) \cite{feng2018wing}, DCFE uses extra $3$D information ($3$D) \cite{valle2018deeply}.

\subsection{Ablation experiments}
\label{sec:ablation_experiments}

We study how different backbones influence the AP and NME in Table \ref{tab:backbones}. As expected, face detection and landmark localization benefit from stronger backbones.

\renewcommand{\arraystretch}{1.5}
	\begin{table}[t]
	\setlength{\tabcolsep}{4.5pt}
	\scriptsize
	\centering
	\begin{tabular}{l|c|c|c|c}
		\hline \noalign{\smallskip}
		Backbone & Easy, AP & Medium, AP & Hard, AP & CED@$0.95$ \\
		\hline
		
		\hline
	  	MobileNetv2 & $0.9597$ & $0.9489$ & $0.8893$ & $3.81$\\
	  	\hline
	  	ResNet-50 & $0.9667$ & $0.9560$ & $0.9011$ & $3.61$\\
	  	\hline
	  	ResNeXt-101 & $0.9674$ & $0.9573$ & $0.9042$ & $3.56$\\
	  	\hline
	  	ResNeXt-152 & $\mathbf{0.9724}$ & $\mathbf{0.9652}$ & $\mathbf{0.9153}$ & $\mathbf{3.39}$\\
	  	\hline

	    \hline
	\end{tabular}
	\caption{Dependence of face detection and landmark localization accuracies on the backbone architectures. AP is calculated on the WIDER FACE \textit{validation} subset. CED@$0.95$ -- a CED value at $95$\% images for 5 facial landmarks on the full AFLW.}
	\label{tab:backbones}
	\end{table}

In Table \ref{tab:ablation} we provide results of ablation experiments. As the baseline we use FPN with the ResNet-50 backbone. Experiments show that face detection and landmark localization benefit from the context modules. Tiny faces (\textit{hard} subset) gain the most. The keypoint head slightly decreases the face detection accuracy indicating about interference between the tasks. Note that this behavior is different from the previously reported results on the multi-task CNNs trained on the COCO dataset \cite{lin2014microsoft} when the additional segmentation head increases detection accuracy \cite{he2017mask,fu2019retinamask}. This means that more advanced architectures or training strategies are needed to get benefits from joint face detection and landmark localization ~\cite{maninis2019attentive,chen2017gradnorm}.

We made experiments to demonstrate dependence of NME on the value of $k_0$ parameter on the AFLW \textit{test} subset. For $k_0=3$ NME is 1.54, for $k_0=4$ NME is 1.57, for $k_0=5$ NME is 1.64. The higher $k_0$ the lower spatial resolution of feature maps used for landmark predictions.

\renewcommand{\arraystretch}{1.5}
	\begin{table}[h]
	\setlength{\tabcolsep}{3pt}
	\scriptsize
	\centering
	\begin{tabular}{l|c|c|c|c}
		\hline \noalign{\smallskip}
		 & Context & Keypoint head & Easy / Medium / Hard, AP & CED@$0.95$\\
		\hline
		
		\hline
	  	\multirow{4}{5em}{ResNet-$50$ + FPN} 
	  	 &   &   & $0.9649$ / $0.9546$ / $0.8980$ & - \\ \cline{2-5}
		 & + &   & $0.9679$ / $0.9580$ / $0.9038$ & - \\ \cline{2-5}
	     &   & + & $0.9631$ / $0.9517$ / $0.8951$ & $3.64$\\ \cline{2-5}
	     & + & + & $\mathbf{0.9667}$ / $\mathbf{0.9560}$ / $\mathbf{0.9011}$ & $\mathbf{3.61}$\\
        \hline 

	    \hline
	\end{tabular}
	\caption{Ablation experiments. AP is calculated on the WIDER FACE \textit{validation} subset. CED@$0.95$ -- a CED value at $95$\% images for $5$ facial landmarks on the \textit{full} AFLW.}
	\label{tab:ablation}
	\end{table}

\renewcommand{\arraystretch}{1.5}
	\begin{table}[h]
	\setlength{\tabcolsep}{4.5pt}
	\scriptsize
	\centering
	\begin{tabular}{l|c|c|c}
		\hline \noalign{\smallskip}
		 & Backbone & Time & GPU type\\
		\hline
		
		\hline
	  	\multirow{2}{7em}{\bf{Our MaskFace}} 
	  	 & ResNet-50  &  20 ms & 2080TI \\
		 & ResNeXt-152 & 55 ms  & 2080TI\\
        \hline 
        \multirow{1}{7em}{RetinaFace \cite{deng2019retinaface}} 
		 & ResNet-152 &  75 ms  & Tesla P40 \\
        \hline 
        \multirow{2}{7em}{RefineFace \cite{zhang2019refineface}} 
	  	 & ResNet-50  &  35 ms & 1080TI \\
		 & ResNet-152 &  57 ms  & 1080TI\\
        \hline 
        \multirow{1}{7em}{DFS \cite{tian2018learning}} 
	  	 & ResNet-50  &  35 ms & Tesla P40 \\
        \hline 

	    \hline
	\end{tabular}
	\caption{Time comparison between different face detection methods for VGA (640x480) images.}
	\label{tab:timings}
	\end{table}

\subsection{Inference time}
\label{sec:inference_time}
\vspace{-0.05cm}
We measure inference time of MaskFace for different backbones using a 2080TI GPU and compare results with recent models. Results are provided in Table \ref{tab:timings}. They show that performance of MaskFace is in line with recent state-of-the-art models. MaskFace spends 0.11 ms to predict landmarks for one face and can achieve about 20 fps for 640x480 images even with the heavy ResNeXt-152 backbone.


\section{Acknowledgments}
The authors thank Stefan Marti, Joey Verbeke, Andrey Filimonov and Vladimir Aleshin for their help in this research project.

\section{Conclusion}

In this paper we have shown that adding the mask head to the face detection models significantly increases localization accuracy of sparse and dense landmarks. The proposed MaskFace model achieves top results in face and landmark detection on several popular datasets. The mask head is very fast, therefore without computational overhead MaskFace can be used in applications with few faces on the scene offering state-of-the-art face and landmark detection accuracies.

\small
\bibliographystyle{ieee_fullname}
\bibliography{MaskFace_arxiv}

\end{document}